\documentclass[10pt,twocolumn,letterpaper]{article}

\usepackage[utf8]{inputenc} % allow utf-8 input
\usepackage[T1]{fontenc}    % use 8-bit T1 fonts
\usepackage{hyperref}       % hyperlinks
\usepackage{url}            % simple URL typesetting
\usepackage{booktabs}       % professional-quality tables
\usepackage{amsfonts}       % blackboard math symbols
\usepackage{nicefrac}       % compact symbols for 1/2, etc.
\usepackage{microtype}      % microtypography
\usepackage{xcolor}         % colors

\usepackage{times}
\usepackage{helvet}
\usepackage{courier}
\usepackage{multicol}
\usepackage{multirow}
\usepackage{lipsum}
\usepackage{graphicx}       % graphics
\usepackage{csquotes}
\usepackage{paralist,hyperref,titlesec,fancyhdr,etoolbox}
\usepackage[ruled, vlined, linesnumbered]{algorithm2e}
\usepackage{amsmath,amssymb,amsfonts}
\usepackage{amsthm}
\usepackage{balance}
\usepackage{booktabs}
\usepackage{float}
\usepackage{flushend}
\usepackage{indentfirst}
\usepackage{makecell}
\usepackage{mathrsfs}
\usepackage{multirow}
\usepackage{nccmath}
\usepackage{setspace}
\usepackage{stmaryrd}
\usepackage{subfigure}
\usepackage{textcomp}
\usepackage{xspace}

\newtheorem{theorem}{Theorem}[]

\newtheorem{proposition}[theorem]{Proposition}

\newcommand{\ie}{\emph{i.e.},\xspace}

\newcommand\figref[1]{Fig.~\ref{#1}}

\newcommand\tabref[1]{Table~\ref{#1}}

\newcommand\secref[1]{Sec.~\ref{#1}}

\newcommand\appref[1]{(App.~\ref{#1})}
\newcommand{\algoname}{LPD\xspace}
\newcommand{\fakeparagraph}[1]
{\vspace{1mm}\noindent\textbf{#1.}}

\title{Understanding and Mitigating the High Computational Cost in \\ Path Data Diffusion}

\author{Dingyuan Shi\footnotemark[2]\;,~Lulu Zhang,~Yongxin Tong,~Ke Xu \\
Beihang University \\
\small \texttt{chnsdy@buaa.edu.cn, zll.zhanglulu@buaa.edu.cn, yxtong@buaa.edu.cn, kexu@buaa.edu.cn}
}

\begin{document}
\maketitle
{
\renewcommand{\thefootnote}{\fnsymbol{footnote}}
\footnotetext[2]{Corresponding Author}
}

\begin{abstract}
Advancements in mobility services, navigation systems, and smart transportation technologies have made it possible to collect large amounts of path data. 
Modeling the distribution of this path data, known as the Path Generation (PG) problem, is crucial for understanding urban mobility patterns and developing intelligent transportation systems.
Recent studies have explored using diffusion models to address the PG problem due to their ability to capture multimodal distributions and support conditional generation.
A recent work devises a diffusion process explicitly in graph space and achieves state-of-the-art performance.
However, this method suffers a high computation cost in terms of both time and memory, which prohibits its application. 
In this paper, we analyze this method both theoretically and experimentally and find that the main culprit of its high computation cost is its explicit design of the diffusion process in graph space.
To improve efficiency, we devise a Latent-space Path Diffusion (\algoname) model, which operates in latent space instead of graph space.
Our \algoname significantly reduces both time and memory costs by up to $82.8\%$ and $83.1\%$, respectively.
Despite these reductions, our approach does not suffer from performance degradation. 
It outperforms the state-of-the-art method in most scenarios by $24.5\%\sim 34.0\%$.
\end{abstract}

\section{Introduction}
\label{sec:intro}

Advancements in mobility services, navigation systems, and smart transportation technologies have led to vast amounts of path data. 
These data have immense potential for a range of smart transportation applications, such as traffic flow prediction \cite{TRPC18Wu}, estimated time of arrival (ETA) calculations \cite{KDD20Hong}, etc. 
To leverage the full potential of this large volume of path data, recent studies have focused on modeling its distribution \cite{NeurIPS21Jain, VLDB23Tian, ICLR24Shi}, \ie to generate paths that align with the distribution of the original data. 

This problem is formally known as the \textbf{Path Generation (PG) problem}, which is defined as follows:
Given a connected undirected graph $G$ with a set of vertices $V$ and edges $E$, a path $p$ is a sequence of vertices \ie $p = \langle v_0, \ldots, v_n \rangle$, where each pair of adjacent vertices $v_i, v_{i+1}$ on the path must be connected by an edge (\ie $(v_i, v_{i+1}) \in E$).
A path dataset $\mathscr{P}$ contains multiple paths.
The unconditional PG problem is: build a sampled distribution $p(\cdot;\theta)$ from which we can generate a set of paths whose distribution resembles the distribution of the given dataset $\mathscr{P}$.
The conditional PG problem is: given an origin $ori$ and a destination $dst$, build a probabilistic distribution $p(\cdot | ori, dst; \theta)$ from which we can sample paths such that the generated paths begin at $ori$ and end at $dst$, while following the distribution of the given dataset $\mathscr{P}$. 
The problem of ``conditional PG'' is often referred to as ``path planning'', and we use these two terms interchangeably.

Methods addressing the PG problem have evolved from hidden Markov models \cite{Ubicomp14Baratchi} to recurrent neural networks (RNNs) \cite{IJCAI17Wu, VLDB22Wang}. 
Over the past decade, transformer-based models \cite{NeurIPS17Vaswani} have gained significant traction due to their success in various fields \cite{NeurIPS20Brown, ICLR21Dosovitskiy}.
Researchers have also applied transformer-based approaches to path data \cite{CIKM22Liang}.

More recently, diffusion models have gained attention for their robust ability to model multimodal distributions and perform conditional generation. 
Some studies have explored the use of diffusion models to capture path distributions for path generation and planning \cite{ICLR24Shi, CVPR23Jiang, CoRR23Zhu}. 
Among them, \cite{ICLR24Shi} was the first to propose a Graph-constrained Path Diffusion model (GPD). 
This model uses a heat conduction-based diffusion process in graph space to incorporate the topological structure, unlocking the potential of diffusion models to address the PG problem. 
The GPD method has achieved state-of-the-art results for both unconditional and conditional path generation.
Specifically, the GPD approach implements a diffusion process in graph space by employing a categorical diffusion mechanism that uses a series of transition probability matrices. 
This design allows the diffusion model to operate within the constraints of graph topology, yielding superior performance compared to previous methods.

However, the GPD method incurs significant time and memory costs, which hampers its practical applications. 
Through careful theoretical and experimental analysis, we found that the main reason for these high computational costs is the \textit{explicit} design of the diffusion process in graph space. Specifically, this explicit design has three key drawbacks.

\begin{itemize}
    \item 
    \textit{High Memory Consumption for Explicit Graph-space Diffusion.} 
    Directly implementing the diffusion process in graph space aids in capturing the graph structure, but it comes with a high memory overhead. 
    Each time step requires a transition probability matrix that defines the transition probability among vertices, resulting in a total memory cost of $O(TV^2)$, where $T$ is the number of diffusion time steps and $V$ is the number of vertices. 
    These matrices are neither sparse nor low-rank, making compression difficult.
    
    \item 
    \textit{High Time Cost for Sampling.}
    The explicit diffusion process applies forward (\ie inject noise) and the reverse (\ie denoise) process to vertex value directly, resulting in the path generation process being tightly coupled with diffusion process.
    Therefore, the sampling cost will be $O(LT)$ where $L$ is path length and $T$ is the number of diffusion steps.
    Even adopting exponential window size strategy, as proposed in GPD, the cost remains high at $O(T\log L)$.

    \item 
    \textit{Extra Computational Cost for Integrating Conditional Information.}
    The explicit design of the diffusion process complicates the integration of conditional information, such as the origin and destination of the paths. 
    To address this, GPD follows the framework outlined in \cite{ICML22Janner}, which employs an auxiliary network to transform the distribution into a conditional one. 
    However, this additional layer introduces extra computational cost.
\end{itemize}

To overcome the above drawbacks associated with high computational costs in path generation, we propose a Latent-space Path Diffusion (\algoname), in contrast to \textit{graph-space} path diffusion.
Our approach transforms the diffusion process into latent space using an encoder-decoder framework.
The latent-space diffusion manner can solve the above three issues at one stroke:
\textit{(i)}, by converting raw data from categorical to numerical, the latent-space diffusion process eliminates the need for complex transition probability matrices, reducing memory costs. 
\textit{(ii)}, the decoupling of the path generation and diffusion processes in latent space makes the time complexity independent of path length. 
\textit{(iii)}, latent-space diffusion facilitates the integration of conditional information, allowing for scale-shifting style modulation of latent embeddings similar to techniques used in text-to-image tasks \cite{CVPR22Rombach, CoRR22Ho}. 

Besides its improvement on efficiency, our \algoname also outperforms GPD in most cases of both conditional and unconditional path generation on effectiveness.
This is likely because the latent space model allows the diffusion process to focus solely on denoising, while path modeling is offloaded to the encoder-decoder structure.
Specifically, experiments on real dataset show that our \algoname has a reduction of $59.5\%\sim 82.8\%$ and $68.1\%\sim 83.1\%$ in terms of time and memory, respectively.
Additionally, our method outperforms the state-of-the-art GPD by $24.5\%$ to $34.0\%$ in terms of path generation performance.

In summary, the contributions of our work are as follows.
\begin{itemize}
    \item
    We scrutinize the current state-of-the-art GPD method and identify the explicit diffusion design as the primary cause of its high computational cost.
    Specifically, we examine the memory cost due to the transition probability matrices, the computational cost of the diffusion process, and the additional computational burden of the auxiliary neural network for conditional generation.
    \item 
    To address these drawbacks, we propose a Latent-space Path Diffusion (\algoname). 
    We first train a variational autoencoder to facilitate conversion between graph space and latent space and then apply a specially designed denoising diffusion probabilistic model to the latent representations of path data.
    \item 
    Our approach significantly mitigates computational cost, with over $80\%$ reductions in memory and time costs, and a performance improvement of $24.5\% \sim 34.0\%$ on path generation tasks.
\end{itemize}

% Also, the latent space diffusion provide a potential for a unified diffusion modeling together with perception data like image or videos.
% We envision our work as an initializer for the adoption of practice experience from these fields.

% The remainder of this paper is organized as follows. 
% In \secref{sec:background}, we provide a brief overview of the GPD method. 
% \secref{sec:analysis} discusses the analysis of its drawbacks. \secref{sec:method} introduces our Latent-space Path Diffusion (\algoname) approach. 
% The experimental results are presented in \secref{sec:exp}.
% \secref{sec:related} covers related work, and we conclude our study in \secref{sec:conclusion}.

\section{Background}
\label{sec:background}
This section provides an overview of the Graph-constrained Path Diffusion (GPD) method \cite{ICLR24Shi}. 
Given an undirected connected graph $G$ with vertices $V$ and edges $E$, GPD models the diffusion process in two stages: (i) diffusion for individual vertices and (ii) diffusion for sequences of vertices (paths).

\fakeparagraph{Diffusion for Individual Vertices}
Following general categorical value diffusion process \cite{NeurIPS21Austin}.
The forward process for individual vertices is defined as below
\begin{equation}
    q(v_t|v_{t-1}) = \text{Cat}(v_t | \mathbf{p}=\mathbf{v}_{t-1}Q_t) 
\end{equation}
The corresponding reverse process is given by:
\begin{equation}
q(v_{t-1}|v_{t}, \hat{v}_0) = \text{Cat}(v_{t-1}| \mathbf{p}\propto \mathbf{v}_tQ^T_t  \odot \mathbf{\hat{v}}_0 \bar{Q}_{t-1})
\end{equation}
where $q(v_{t}|v_{t-1})$ is a categorical random variable following the probability $\mathbf{p}$.
The probability $\mathbf{p}$ is denoted as a row vector and $Q_t$ is the transition probability matrix.
$Q_t[i,j]$ is the probability transition from categorical $i$ to $j$, and $\bar{Q}_t=\prod_{i=1}^t Q_i$

The transition probability $Q_t$ is designed to reflect the graph's adjacency structure, inspired by heat conduction in graph space. 
It is defined as below.
\begin{equation}
\label{equ:qheat}
    Q_t = \exp\{(\mathbf{A}-\mathbf{D})t\}
\end{equation}
where $\mathbf{A}$ and $\mathbf{D}$ are adjacent matrix and degree matrix, respectively.
The adjacent matrix contains 0/1 values, where the value 1 at the $i$-th row and the $j$-th column denotes an edge between vertex $i$ and $j$.
\ie $\mathbf{A}[i,j]$ is 1 if $(i,j)\in E$ else is 0.
$\mathbf{D}$ is a diagonal matrix and $\mathbf{D}[i,i]$ is the degree of the $i$-th vertex.
This matrix captures the graph structure and $\bar{Q}_t$ can be quickly computed due to the property of matrix exponential.

\fakeparagraph{Diffusion for Sequences of vertices}
For a sequence of vertices (\ie path) $\mathbf{x}$, maintaining the connectivity constraints is hard during diffusion process, so GPD simply extends it from Cartesian product of vertices.
\begin{equation}
    q(\mathbf{x}_t|\mathbf{x}_0) = \Pi_{i=1}^{|\mathbf{x}|} q(\mathbf{x}_t^i|\mathbf{x}_0^i) = \otimes_{i=1}^{|\mathbf{x}|} \text{Cat}(\mathbf{x}^i_t|\mathbf{p}=\mathbf{x}^i_0{\bar{Q}}_t)
\end{equation}

Accordingly, the reverse process is as below.
\begin{equation}
\label{equ:reverse-path}
    q(\mathbf{x}_{t-1}|\mathbf{x}_t, \mathbf{\hat{x}}_0) = \otimes_{i=1}^{|\mathbf{x}|} \text{Cat}(\mathbf{x}^i_{t-1}| \mathbf{p}\propto \mathbf{x}^i_t Q^T_t  \odot \mathbf{\hat{x}}^i_0 \bar{Q}_{t-1})
\end{equation}

\section{Understanding the High Computational Costs of the GPD Method}
\label{sec:analysis}

The primary reason for GPD's low efficiency lies in its explicit design for the diffusion model in graph space. 
This approach has three significant drawbacks, which we will detail below.

\subsection{High Memory Consumption for Explicit Graph-space Diffusion}

\begin{figure*}[tbp]
\begin{minipage}[]{0.6\textwidth}
\subfigure[Graph Structure]{
    \includegraphics[width=0.4\textwidth]{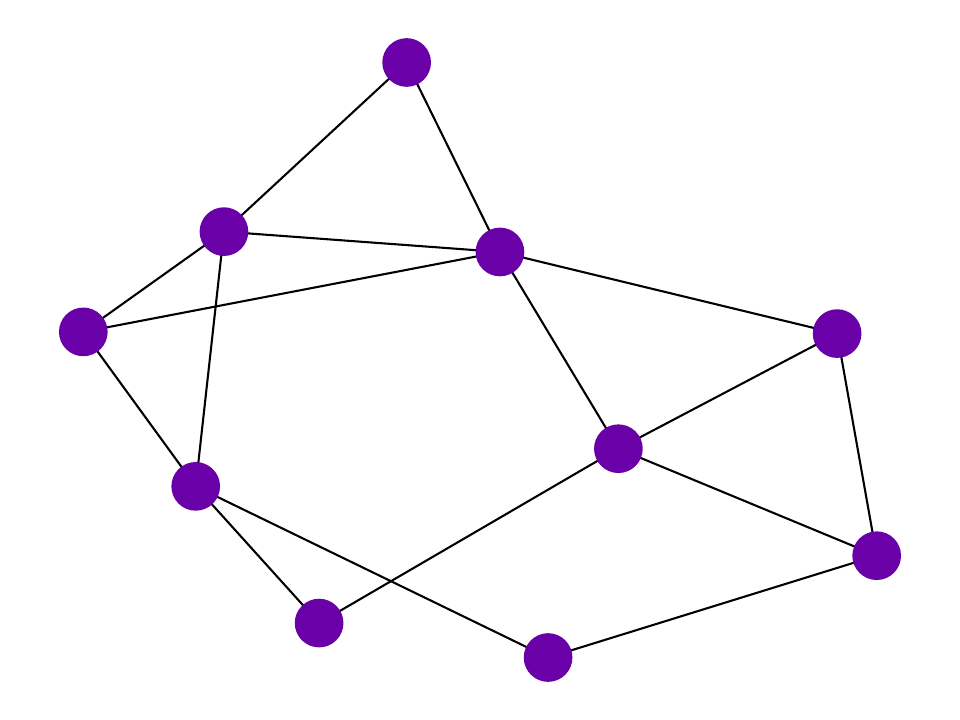}
    \label{subfig:graph}
    % \Description{A graph with 10 vertices.}
}
\quad
\subfigure[Matrix Values]{
    \includegraphics[width=0.5\textwidth]{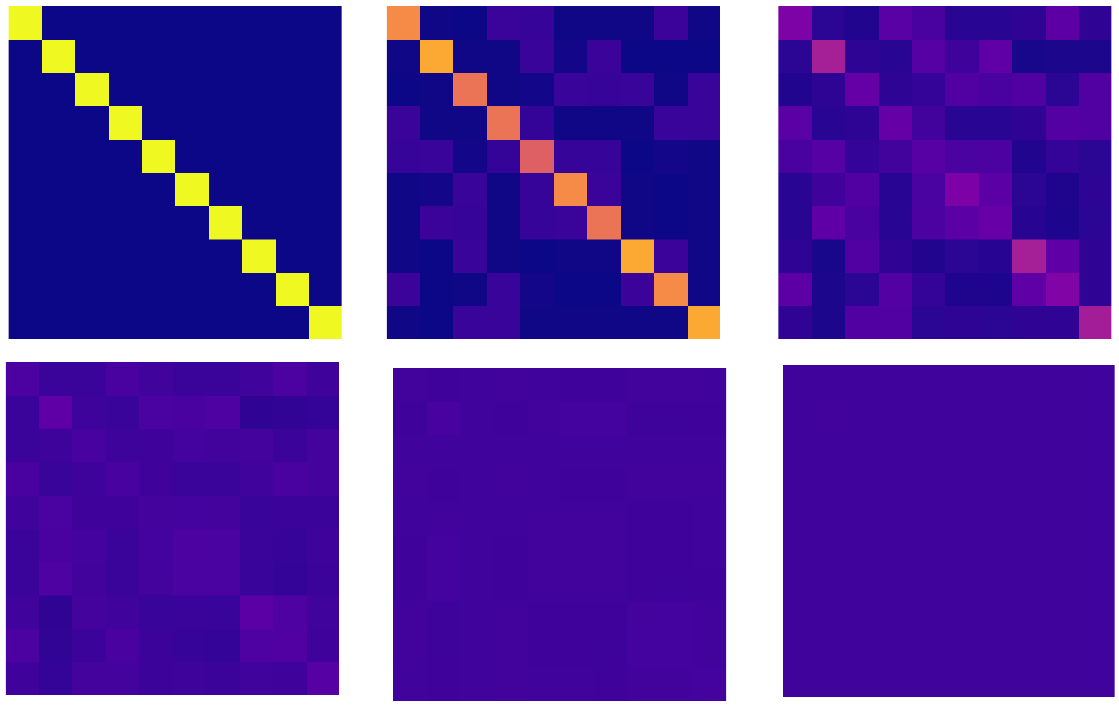}
    \label{subfig:mat}
    % \Description{Six matrices indicating value change.}
}
\caption{A toy example for Transition Probability Matrix (TPM) in GPD method.}
\label{fig:tpm}
% \Description{Two polylines.}
\end{minipage}
\quad
\begin{minipage}[]{0.3\textwidth}
\centering
\includegraphics[width=0.9\textwidth]{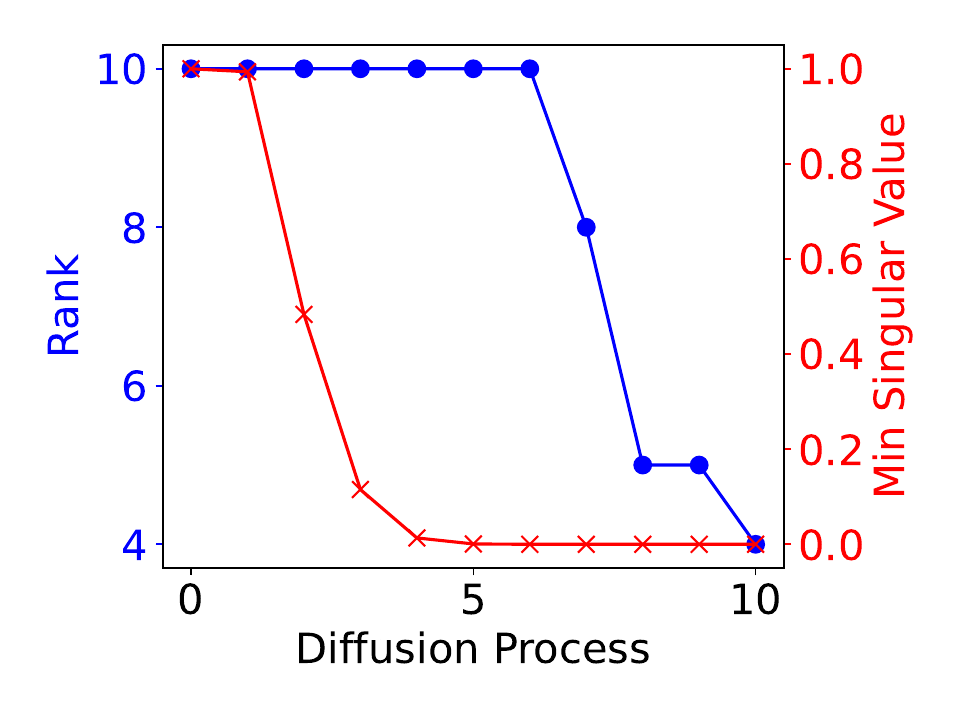}
\caption{Rank and singular values change.}
% \Description{Two poly lines for rank and sigular value change.}
\label{fig:rank}
\end{minipage}
\end{figure*}

Due to its explicit design, the GPD method requires a transition probability matrix (TPM) $Q_t$ at each time step. 
This matrix is defined as in \eqref{equ:qheat}. 
Consequently, the total memory consumption is $O(TV^2)$. 
Given 1000 time steps and a graph with 2000 vertices, the memory cost can reach magnitudes of up to $O(10^9)$, resulting in substantial memory consumption overhead. 
The problem is exacerbated by the fact that these matrices are difficult to compress due to two key properties.
\textit{(i) Lack of Sparsity.} As $t$ increases, the transition probability matrix $Q_t$ becomes progressively less sparse.
\textit{(ii) High Rank.} The rank of $Q_t$ tends to be high, further complicating any attempts to compress the matrices.

These properties contribute significantly to the computational and memory inefficiencies of the GPD method, limiting its scalability and practical application.
Here we provide intuitive explanations from the perspective of its corresponding physical meaning.
The detailed analysis is put in \appref{app:proof}.

\fakeparagraph{Lack of sparsity}
The TPM $Q_t$ represents the proportion of heat transferred between vertices.
Specifically, the value in the $i$-th row and $j$-th column indicates the total heat conducted from vertex $i$ to vertex $j$.
Initially, when no heat conduction has occurred, $Q_t$ is an identity matrix, as all heat is retained by its original vertex. 
As $t$ increases, heat begins to diffuse across the graph, resulting in non-zero values in the TPM as heat is transferred between connected vertices. 
This process makes $Q_t$ progressively denser, leading to a lack of sparsity.

\fakeparagraph{High rank}
Heat conduction follows the second law of thermodynamics, indicating that heat will naturally flow from a source vertex to other connected vertices without an increase in the heat of the source. 
This is reflected in the TPM, where each row represents the distribution of heat from a specific source vertex. 
Since heat always transfers outwards, the diagonal elements of the TPM tend to have the largest values, as they represent the original source of heat.
Given that the sum of each row in the TPM is 1, these large diagonal values imply that each row is linearly independent, leading to a high rank in the matrix. 
Only when $t$ approaches infinity do all rows converge to the same value $1/|V|$, reducing the rank to 1. 
During this process, the singular value becomes smaller, which could also lead to some potential numerical instabilities.

Below we provide a toy example for better illustration of the above two properties.
The theoretical analysis is put in \appref{app:proof}.

Here we provide a toy example for illustration.
Consider a small graph with 10 vertices whose structure is shown in \figref{subfig:graph}.
The TPM values for different diffusion timesteps $Q_t$ are depicted in \figref{subfig:mat}, wherein cooler color indicates lower values.
At $t = 0$, all heat is concentrated at their source, resulting in $Q_0$ being an identity matrix.
As $t$ increases, the heat start to diffuse to other vertices, eventually becoming nearly uniform.
This demonstrates that as $t$ increases, the TPM transforms from sparse to dense.
\figref{fig:rank} shows how the rank and the minimum singular value of the TPM change as $t$ increases.
The rank of $Q_t$ drops from 10 to 4 as diffusion progresses. Even when the matrix becomes almost uniform, it still retains a significant number of ranks, indicating a high degree of linear independence among rows. 
This structure reflects the ongoing heat diffusion across the graph. 
However, despite this drop in rank, the minimum singular value becomes very small, suggesting a high potential for numerical instability. 
This can lead to large values in the inverse matrix, which in turn may cause computational issues in practice.

% \begin{figure}[htbp]
% \centering
% \includegraphics[width=0.5\textwidth]{figures/memo_cost.pdf}
% \caption{Memory Cost}
% \label{fig:memo_cost}
% \end{figure}

% Next we test the memory consumption of GPD method under real-world dataset.
% When batch size is set to 1, the memory usage is already as high as $6,535$ MiB.
% As the batch size increases to 512, the memory cost soars to $35,531 $ MiB, which is a quite large memory load.
% This considerable memory load underscores the high computational demands of the GPD method, particularly when processing large-scale data.
The experiment results in \secref{subsec:expefficiency} clearly show that the GPD method's reliance on large TPM leads to significant memory requirements. 

\begin{figure*}[tbp]
\begin{minipage}[]{0.3\textwidth}
\includegraphics[width=1.0\textwidth]{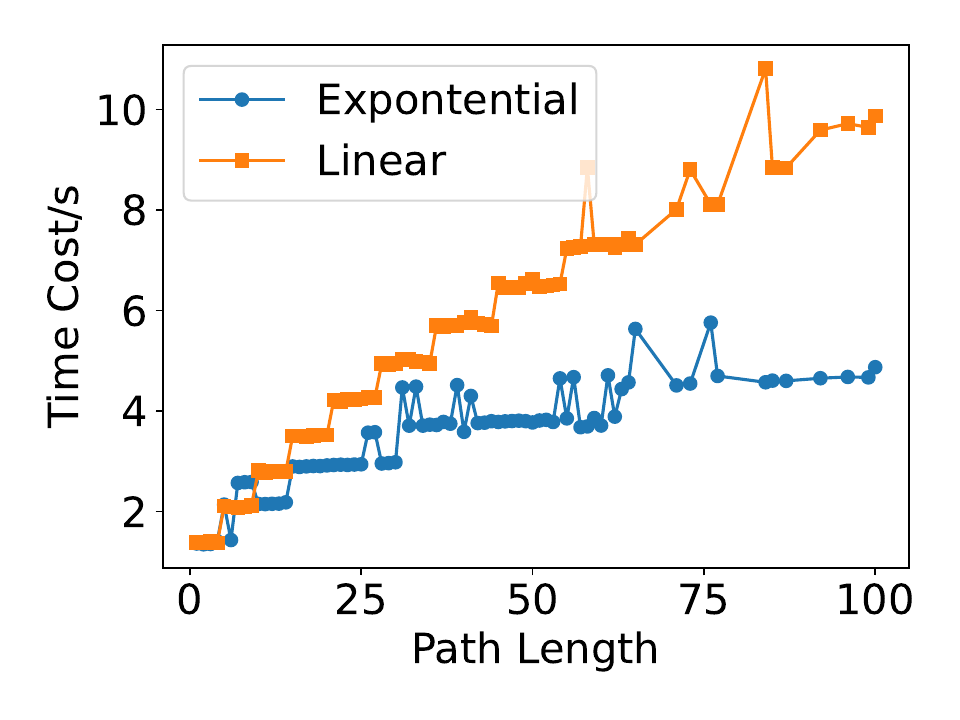} 
\caption{Time comparison between linear/exponential strategy.}
% \Description{Two polylines.}
\label{fig:time_cmp}   
\end{minipage}
\quad
\begin{minipage}[]{0.65\textwidth}
\subfigure[Exponential]{
    \includegraphics[width=0.45\textwidth]{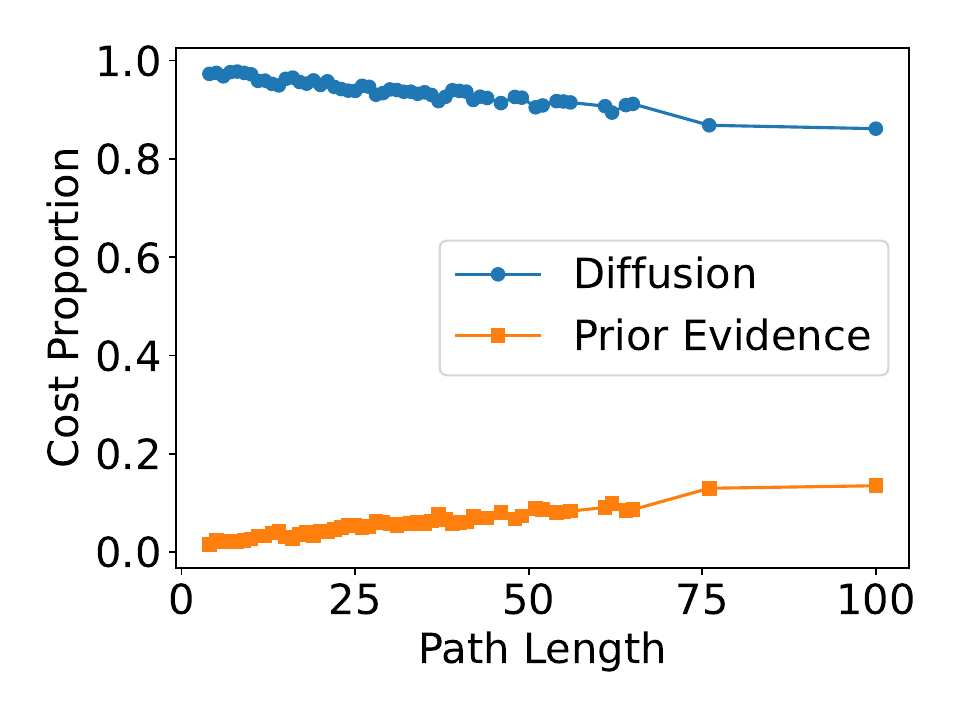}
    \label{subfig:breakdown_exp}
    % \Description{Two polylines.}
}
\subfigure[Linear]{
    \includegraphics[width=0.45\textwidth]{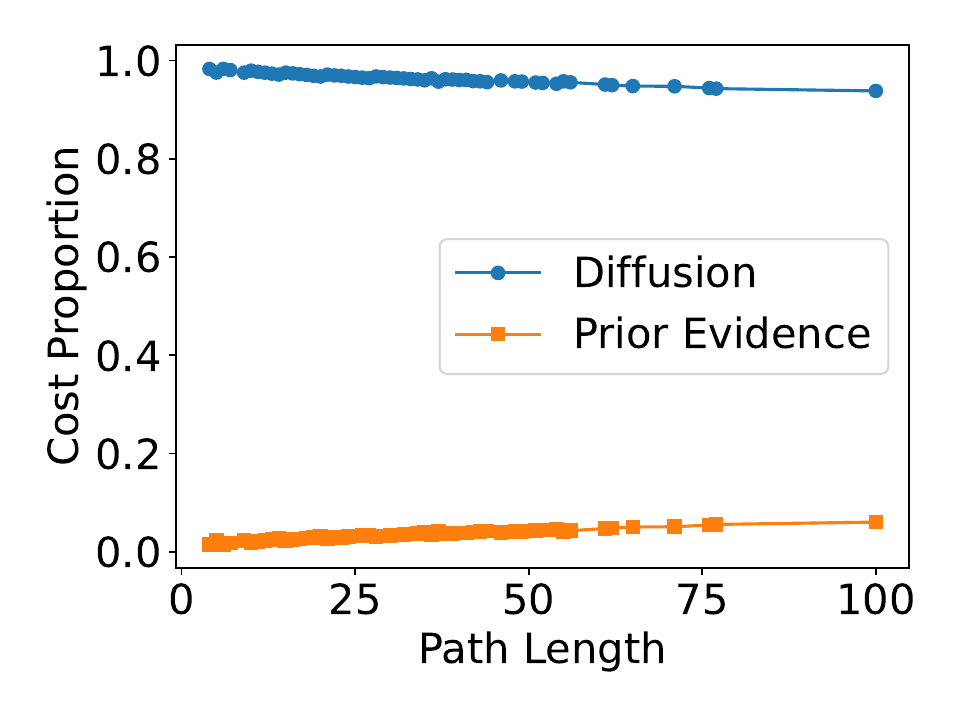}
    \label{subfig:breakdown_lin}
    % \Description{Two polylines.}
}
\caption{Time cost breakdown.}
\label{fig:tc_breakdown}
\end{minipage}
\end{figure*}

\subsection{High Time Cost Caused by Sampling}
The sampling process incurs significant computational costs due to the coupling between path generation and the diffusion process, a consequence of the explicit design of diffusion in graph space. 

The exact length of paths is often unknown in advance, especially in planning scenarios, leading to a process that operates in an auto-regressive manner. 
So the generation process forms a below cycle: \textit{(i)} concatenate noise to the current path prefix \textit{(ii)} denoise path using diffusion process \textit{(iii)} continue this cycle till we hit the destination or reach the max length.
The most naive method is to concatenate noise with a length of 1 once at a time, which results in diffusion times linear to the path length.
We call this \textit{linear strategy}.

GPD method adopts an \textit{exponential strategy}, where the length of the noise concatenated is equal to the current prefix at each step.
While this approach reduces time complexity compared to the linear strategy, it still results in high computational costs.

We test the time cost of both strategies (see \figref{fig:time_cmp}).
As the path length increases, the time cost for the linear strategy grows linearly, while the exponential strategy's cost rises logarithmically. 
While the exponential strategy is more efficient, particularly for longer paths, the computational cost can still be high. 
Notably, when the path length reaches 30, the required time exceeds 4 seconds, which will result in intolerable latency in daily applications.
In navigation applications, users can get the planning scheme within nearly 1-3 seconds, highlighting the need for more efficient path generation methods.

\subsection{Extra Computational Cost for Integrating Conditional Information}
In this subsection we demonstrate that how the integration of conditional information incurs extra computational cost.
We broke down the time consumption to assess the proportion of time taken by each component. 
\figref{fig:tc_breakdown} shows that the diffusion process accounts for nearly 90\% of the total time cost. 
Additionally, prior evidence, which is used for GPD method for conditional distribution calculation, also contributes approximately 10\% to the time cost.
As the path length increases, the proportion of time dedicated to prior evidence calculation also increases, consistent with the auto-regressive approach. 
For the exponential strategy, the proportion can reach 20\%, while for the linear strategy, it tends to stabilize around 10\%.

Although the exponential strategy reduces the overall time cost, it also highlights the extra overhead caused by prior evidence calculation. 
This suggests that the computation required for conditional generation is a notable source of inefficiency, and its impact becomes more pronounced with the use of the exponential strategy.

These findings indicate that the prior evidence calculation introduces a non-trivial computational overhead that should not be overlooked in the design of more efficient path generation methods.

Based on the above analysis, the key factor contributing to the low efficiency of the GPD method is its explicit design of the diffusion process in graph space. 
Consequently, a promising approach to improving efficiency is to transition from diffusion in graph space to diffusion in latent space, as we will describe in the following section.

\section{Latent-space Path Diffusion for Efficiency Improvement}
\label{sec:method}
In this section, we introduce our Latent-space Path Diffusion (\algoname), which comprises two key components: \textit{(i)} an encoder-decoder architecture for converting between latent and graph space, and \textit{(ii)} a diffusion process that operates in latent space. 
These components work together to improve computational efficiency and reduce memory consumption while generating paths that adhere to the underlying graph structure.

\subsection{Encoder-Decoder for Path Data}

To enable a diffusion process in latent space for path data, we build an encoder-decoder framework to convert path data between graph space and latent space, similar to the latent diffusion model for images \cite{CoRR22Ho}. 
This framework is based on a Variational Autoencoder (VAE) \cite{ICLR14Kingma} with transformer-based components.
The encoder is a bidirectional transformer, while the decoder is a causal transformer designed for auto-regressive generation. 

The encoder, denoted by $\mathscr{E}$, converts a path of length $L$ into a fixed-size latent representation with length 2. 
Formally, the encoder maps data from graph space to latent space, $\mathscr{E}: \mathbb{R}^{L \times V} \rightarrow \mathbb{R}^{2 \times C}$, where $V$ is the number of vertices, and each vertex in a path is represented as a one-hot vector of length $V$. The channel size $C$ determines the dimensions of the latent space.

The two-element output from the encoder contains the embedding of the $[CLS]$ token and the mean embedding derived from all tokens after the bidirectional self-attention mechanism. 
The $[CLS]$ token is added at the beginning of each path, following BERT \cite{NAACL19Devlin}.
Additionally, we take the mean of the latent embedding for the entire paths for better representation.

The decoder $\mathscr{D}$, samples hidden features from the encoder and then generates a path in an auto-regressive manner. 
It uses the two tokens from the encoder as a prefix and then proceeds with generation. 
Formally, the decoder maps from latent space to graph space, $\mathscr{D}: \mathbb{R}^{2 \times C} \rightarrow \mathbb{R}^{L \times V}$.

Our latent space representation significantly reduces the representation size from $L \times V$ to $2 \times C$, where $L$ is the path length, $V$ is the number of vertices, and $C$ is the channel size. 
This reduction in representation size has two key benefits: \textit{(i)} it significantly reduces memory and time costs, and \textit{(ii)} it decouples the latent space representation from path length, facilitating more flexible diffusion processes in latent space.

\subsection{Path Diffusion in Latent Space}
The trained encoder and decoder allow us to convert paths into latent space, ensuring that the latent distribution resembles a Gaussian distribution, and then convert them back to the original data space.
In this subsection, we explain how we design the diffusion process in this latent space.

By embedding the raw data into a smaller Euclidean space, we can directly use a diffusion model for path generation, without the need to design complex diffusion processes specific to graph space like in GPD. 
We adopt a Denoising Diffusion Probabilistic Model (DDPM) \cite{NeurIPS20Ho}, which is suitable for our latent space because the encoder, guided by KL divergence loss, creates a distribution with restricted variance. 
This characteristic allows the diffusion model to work efficiently by converting the latent space into a tractable Gaussian distribution.
The forward and backward diffusion processes are defined as follows.

\begin{equation}
    \mathbf{z_t} = \alpha_{t - 1} \mathbf{z}_{t-1} + \beta_{t-1}\mathbf{\epsilon}_t
\end{equation}

\begin{equation}
    \hat{\mathbf{z}}_{t-1} = \frac{1}{\sqrt{\alpha_t}}(\mathbf{z}_t - \frac{1-\alpha_t}{\sqrt{1 - \bar{\alpha}_t}}\mathbf{\epsilon}_{\psi}(\mathbf{z}_t, t)) + \sigma_t \mathbf{\epsilon}_t, \forall t=1,.., T
\end{equation}
where $\epsilon_t$ denotes a standard Gaussian noise, $\alpha_t$ and $\beta_t$ compose the noise scheduler that controls the signal-noise-ratio.
Our diffusion model $\epsilon_\psi$ is designed to predict the noise in the latent space.

Regarding the network architecture, we use an off-the-shelf Unet implementation designed for 2D images, adapting it to work with 1D data by compressing the height and width dimensions ($H×W$) into a single sequence length dimension ($L$).
This approach has proven to be effective in our experiments.

\begin{figure*}[h]
    \centering
    \includegraphics[width=0.8\textwidth]{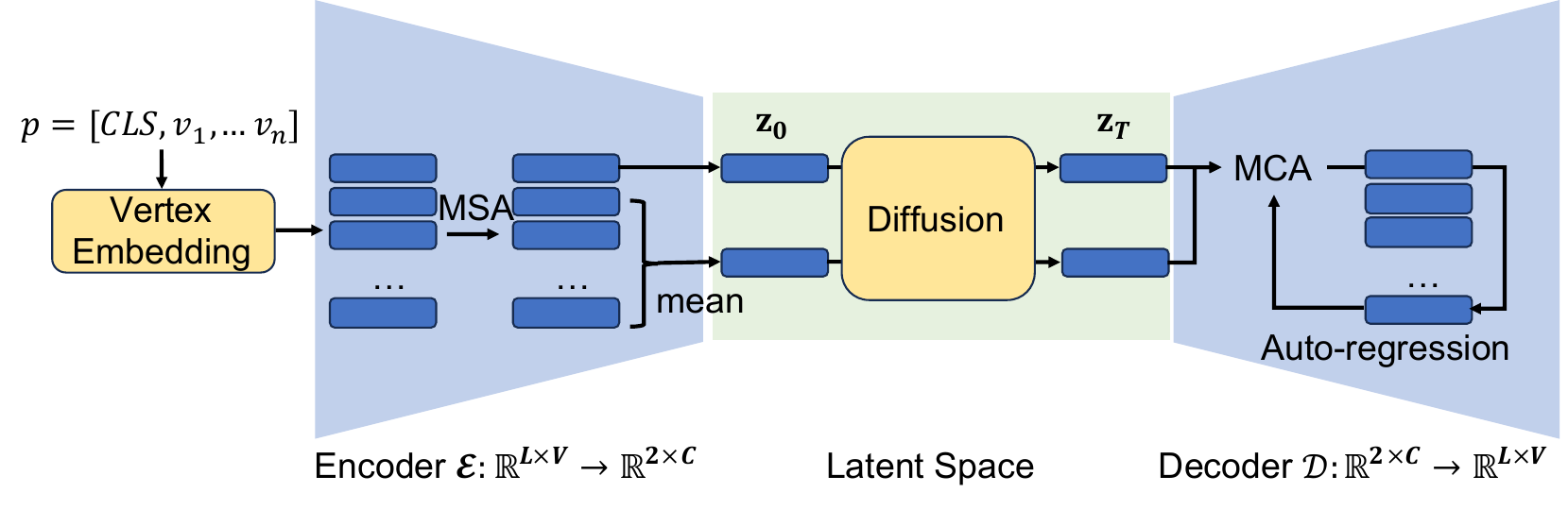}
    \caption{Our encoder-decoder framework.}
    % \Description{Framework.}
    \label{fig:framework}
\end{figure*}

Our encoder-decoder framework are shown in \figref{fig:framework}.
% \fakeparagraph{Train encoder and decoder in a VAE-style}
We train the encoder and decoder in a VAE-style framework, incorporating reconstruction loss and KL-divergence. 
The reconstruction loss involves minimizing the cross-entropy between the ground truth and the decoder's output logits. 
The KL-divergence term penalizes differences between the encoder's output distribution and a Gaussian distribution. 
The overall training loss is given as below.

\begin{equation}
\begin{aligned}
\min_{\theta, \phi}\mathscr{L} &= \mathscr{L}_{recon} + \lambda \mathscr{L}_{KL} \\
&= \mathbb{E}_{x\sim data, z\sim \mathscr{E}_\theta(x)}[CE(\mathscr{D}_\phi(z), x) + \lambda KL(z||\mathscr{N})]
\end{aligned}
\end{equation}
where $\mathscr{N}$ denotes the standard normal distribution.

\begin{figure*}
    \centering
    \includegraphics[width=0.8\textwidth]{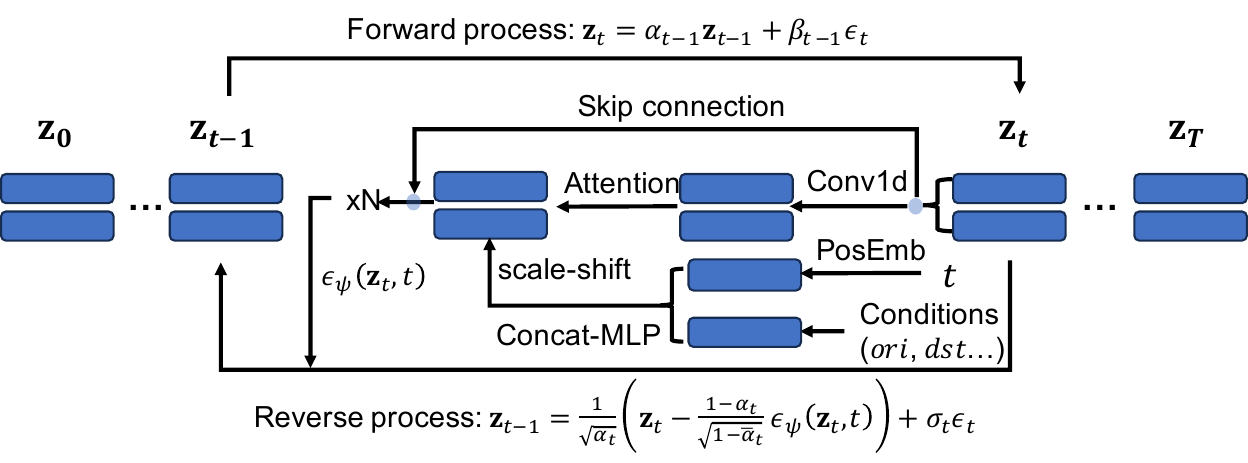}
    \caption{Latent diffusion model.}
    % \Description{Latent diffusion.}
    \label{fig:diffusion}
\end{figure*}

The diffusion process is shown illustrated in \figref{fig:diffusion}.
% \fakeparagraph{Add origin and destination as condition}
To integrate conditions into diffusion models, we extract the embedding of the origin and destination from the encoder $\mathscr{E}_\theta$, then we concatenate them together and feed them into a multi-layer perception (MLP) to convert the embedding to the same dimension as the diffusion time embedding.
For the U-net structure, we integrate the origin-destination condition into the time step condition using an MLP, based on which we calculate the scale $w$ and shift $s$ to modulate the original latent value by $\mathbf{z}_{cond} = (1 + w)\mathbf{z} + s$.
When training the conditional diffusion model, we freeze the encoder and decoder to reduce training overhead.

\section{Experiment}
\label{sec:exp}
\subsection{Experiment Settings}

\fakeparagraph{Datasets}
We use the Didi GAIA open path datasets  \footnote{\url{https://gaia.didichuxing.com}}.
We choose two datasets from two cities for validation.
The raw data is provided in longitude-latitude format. 
We conduct map matching \cite{ECDA18Meert} to bind the coordinates to road network which can be fetched on open street map \footnote{\url{https://www.openstreetmap.org}}.
Hence we can get a city road network in graph and the paths as sequences of vertices.
We choose two cities whose road networks are represented as graphs with 2717 and 2195 vertices, respectively.
The corresponding number of paths are 118,535/ 89,339 with average length of 26.76 / 28.61.

\fakeparagraph{Baselines}
For unconditional path generation, we choose CSSRNN (\cite{IJCAI17Wu}) and MTNet (\cite{VLDB22Wang}) methods.
Both methods adopt RNN nerual network.
MTNet was the state-of-the-art method before GPD.
For conditional path generation (\ie path planning), we choose NMLR (\cite{NeurIPS21Jain}) and KS (\cite{VLDB23Tian}) method.
NMLR employs a neural network to predict the probability of choosing each edge, then conduct graph search method to maximum the likelihood and KS improves it by adding a relay vertex prediction to reduce the accumulation error of long sequences.
We choose state-of-the-art method GPD \cite{ICLR24Shi} for both unconditional and conditional generation tasks for comparison.

\fakeparagraph{Evaluation Metrics}
For effectiveness evaluation, the key idea is to measure the similarity between generated and ground truth paths.
There are plenty of manners to evaluate the similarity between paths, following the convention of previous work \cite{VLDB22Wang, ICLR24Shi}, we use Neg-Log-Likelihood (NLL), Edit Distance with Real cost (EDR) and Longest Common Subsequence (LCS) for evaluation.
The definition of NLL is as follows.
\begin{equation}
    NLL(\mathbf{x},\theta) = -\sum_{j=1}^{|\mathbf{x}|-1}\log p_{\theta}(v_{j+1}|v_1,...,v_j) 
\end{equation}
The LCS and EDR can be calculated via a classical dynamic programming strategy.
The NLL can help measure the distribution similarity, while LCS and EDR evaluate the path similarity with regard to graph structure and geographical space, respectively.

To assess the robustness of our method, we use the Beat Ratio (BR), which measures the number of cases where our approach outperforms the simplest baseline methods. 
We choose this metric because it is not reasonable to use variance as an indicator of robustness, given that the path lengths in our dataset exhibit significant variability, leading to large variances in other metrics. 
By focusing on the Beat Ratio, we can evaluate the consistency of our method's performance across a wide range of scenarios.

We use time cost and memory cost for efficiency evaluation.

\fakeparagraph{Implementaion Details}
The embedding dimension for each vertex is set to 256.
The encoder and decoder layers both have three layers and 8 attention heads.
The diffusion model has 512 latent dimension and 4 resnet blocks.
We adopt Gaussian diffusion process with DDPM noise scheduler, the diffusion process has 50 timesteps.
For training process, both the encoder-decoder and diffusion models are trained with batch size 128 in three epochs, the learning rate are fixed at 0.005.
Our source code can be found at \url{https://anonymous.4open.science/r/Latent-space-Path-Diffusion-FD86/}.
Each model is trained using GPU A100-40G within 6 hours.

\subsection{Efficiency Evaluation}
\label{subsec:expefficiency}
\begin{figure}[htb]
\centering
\subfigure[Unconditional Generation]{
    \includegraphics[width=0.22\textwidth]{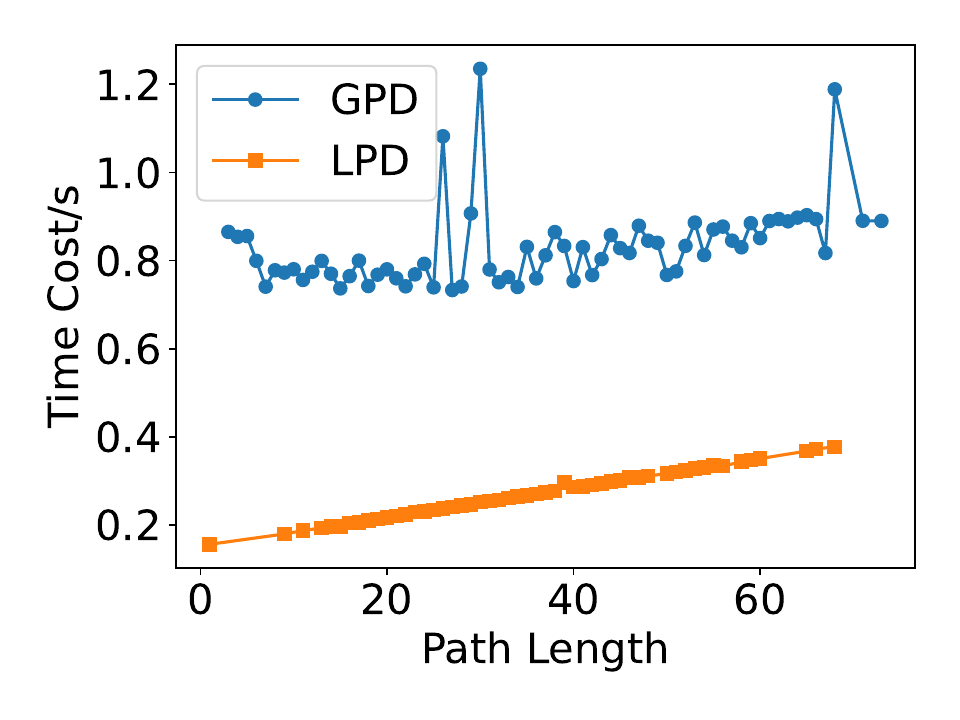}
    \label{subfig:time_cmp_gen}
    % \Description{Two polylines.}
}
\subfigure[Conditional Generation]{
    \includegraphics[width=0.22\textwidth]{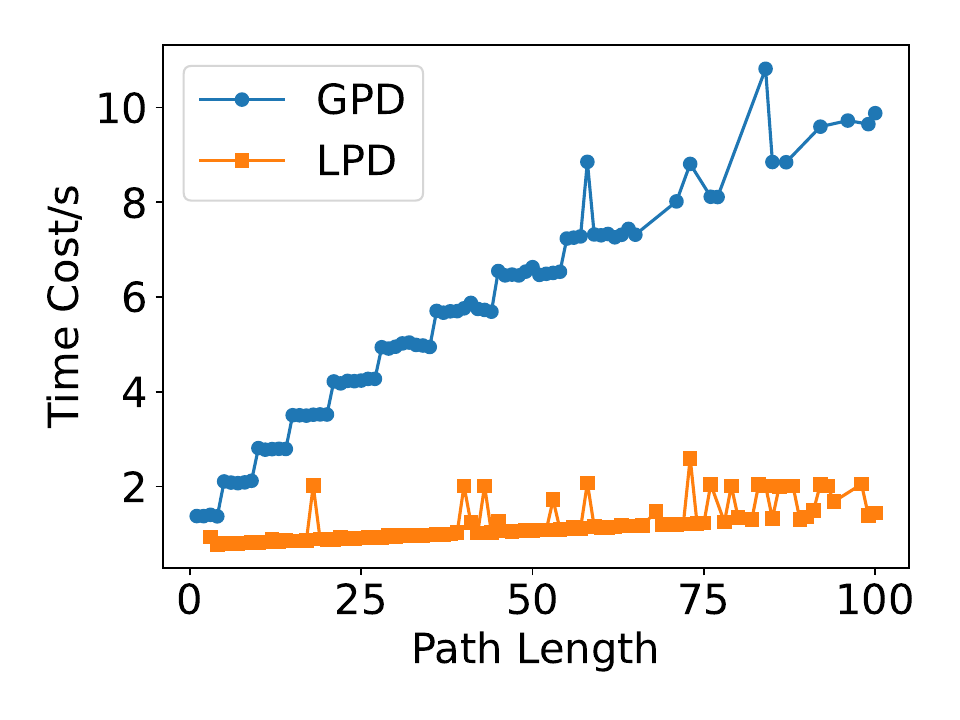}
    \label{subfig:time_cmp_plan}
    % \Description{Two polylines.}
}
\caption{Inference time cost comparison}
\label{fig:xa_plan_time}
\end{figure}

\fakeparagraph{Time efficiency}
The time cost comparison for both unconditional and conditional generation are shown in \figref{subfig:time_cmp_gen} and \figref{subfig:time_cmp_plan}, respectively.
Overall, our method can significantly reduce time cost by up to $78.3\%$ and $82.8$ in terms of unconditional and conditional generation, respectively.

For unconditional path generation, our \algoname can beat the GPD method by reducing time from $67.8\%\sim 78.3\%$.
The GPD method adopts a Unet structure, whose CNN-style generation results in a slow increase in time cost as the path length increases.
The time cost of our \algoname increases due to the auto-regressive decode manner.
Despite this difference, our method is generally faster than the GPD method in terms of all path lengths. 
This advantage comes from the efficiency gained by operating in latent space, which substantially reduces computational overhead and leads to faster path generation compared to GPD's approach.

For conditional path generation, our \algoname has a significant advantage due to its diffusion process occurring in latent space. 
This characteristic ensures that the diffusion process is independent of the path length, leading to a slow increase in time cost as path length grows.
In contrast, even though the GPD method uses an exponential strategy (as mentioned in Section \ref{sec:analysis}), its time cost still increases logarithmically. 
When the path length reaches 50, the time cost with GPD exceeds 4 seconds, which can be impractical for real-time applications. 
This limitation underscores the benefit of diffusion in latent space, as it provides a more scalable and efficient approach to conditional path generation.
Concretely, \algoname reduces conditional path generation time cost by $59.5\%\sim 82.8\%$.

\begin{figure}[htbp]
\centering
\subfigure[Time cost breakdown]{
\includegraphics[width=0.22\textwidth]{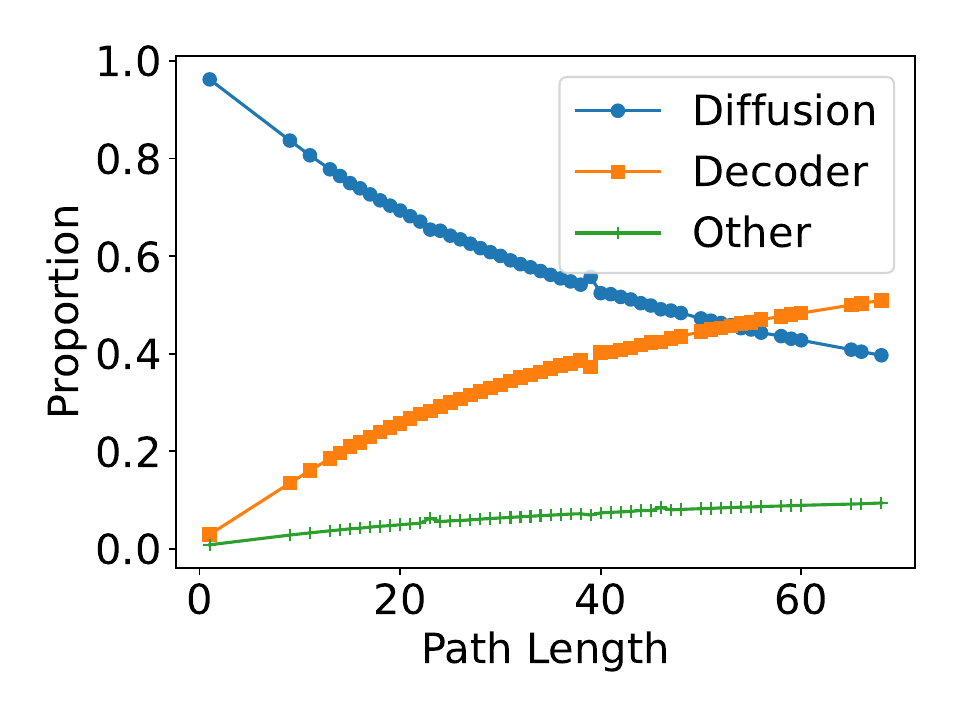}
\label{fig:time_cmp_break}
% \Description{Three polylines.}
}
\subfigure[Memory cost comparison]{
\includegraphics[width=0.22\textwidth]{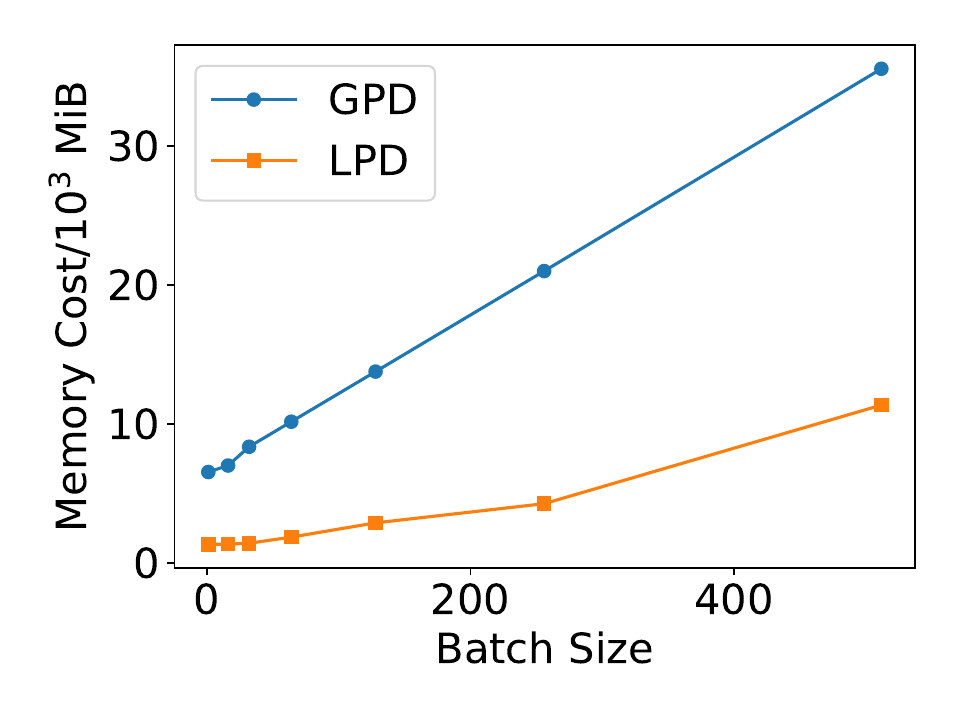}
\label{fig:memo_cost_cmp}
% \Description{Two polylines.}
}
\caption{Cost Analysis}
\end{figure}

\figref{fig:time_cmp_break} provides a breakdown of the time consumption in our method. 
The results show that as path length increases, the proportion of time attributed to diffusion decreases, while the time cost associated with the decoder increases. 
This pattern suggests that the slight increase in time cost for our method is due to the auto-regressive process in the decoding phase.
However, the diffusion process in latent space maintains a consistent time cost, and its proportion relative to the total cost decreases as path length grows. 
This observation aligns with the design advantages of latent-space diffusion, where the diffusion process remains stable despite variations in path length, emphasizing the efficiency and scalability of our approach.

\fakeparagraph{Memory efficiency}
Latent space path diffusion also helps reduce memory costs, aligning with earlier experiences in image generation. 
However, compared to latent diffusion in image generation, the reasons for our memory reduction are twofold: 
\textit{(i)} Similar to image generation, diffusion in a latent space significantly reduces time and memory costs. 
In our \algoname, the latent space contracts from the original path length to 2, thereby reducing the diffusion space from $O(\mathbb{R}^{L \times V})$ to $O(\mathbb{R}^{2 \times C})$, where $2 < L$ and $C < V$. 
This compression greatly diminishes the memory requirements.
\textit{(ii)} Particularly, by performing diffusion in latent space, our diffusion process is no longer tied to the graph structure, eliminating the need for the transition probability matrix. 
This removes the memory overhead associated with the matrix, which can consume up to $O(T \times V^2)$ in terms of memory.

These factors contribute to a significant reduction in memory cost. Experimental results demonstrate that our approach reduces memory consumption by $68.1\% \sim 83.1\%$.

\begin{figure*}[t]
\centering
\subfigure[Real]{
    \includegraphics[width=0.23\textwidth]{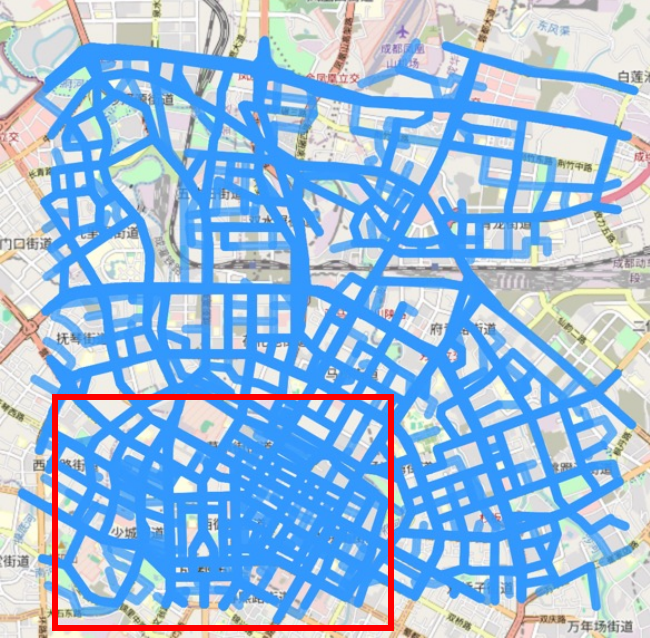}
    \label{subfig:cd_gen_real}
    % \Description{Map with routes.}
}
\quad
\subfigure[LPD (Ours)]{
    \includegraphics[width=0.24\textwidth]{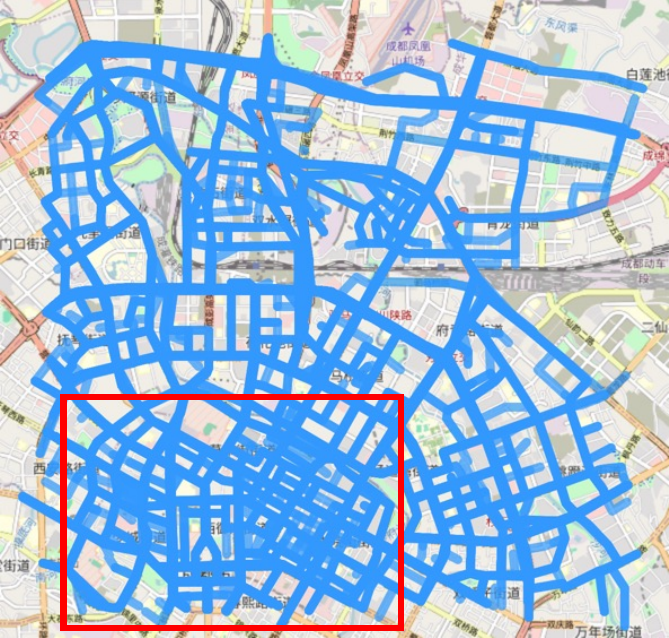}
    \label{subfig:cd_gen_lpd}
    % \Description{Map with routes.}
}
\quad
\subfigure[GPD]{
    \includegraphics[width=0.24\textwidth]{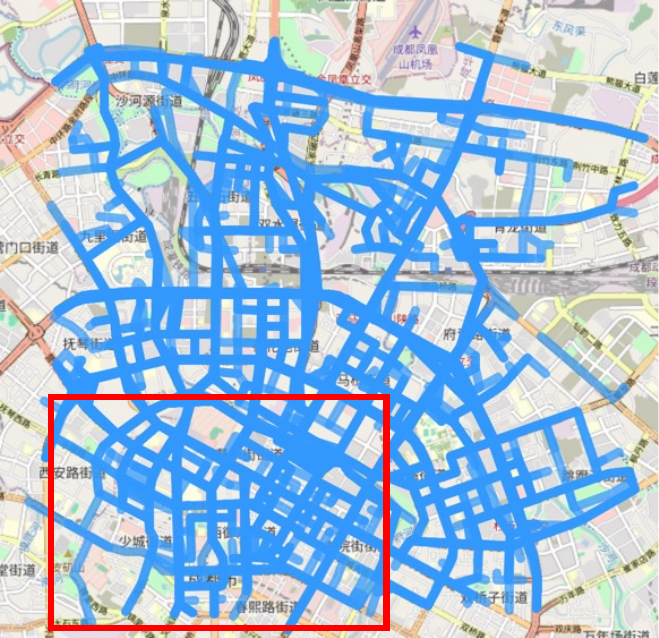}
    \label{subfig:cd_gen_gpd}
    % \Description{Map with routes.}
}
\caption{City A Generated Path Comparison.}
\label{fig:cd_gen}
\end{figure*}

\begin{figure*}[htb]
\centering
\subfigure[Real]{
    \includegraphics[width=0.24\textwidth]{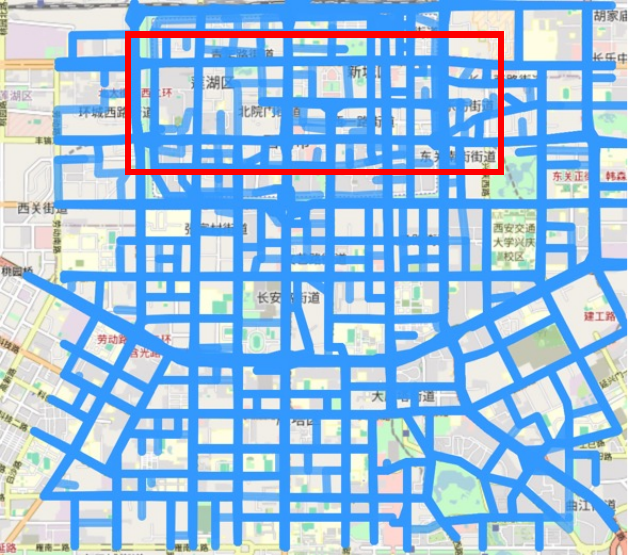}
    \label{subfig:xa_gen_real}
    % \Description{Map with routes.}
}
\quad
\subfigure[LPD (Ours)]{
    \includegraphics[width=0.24\textwidth]{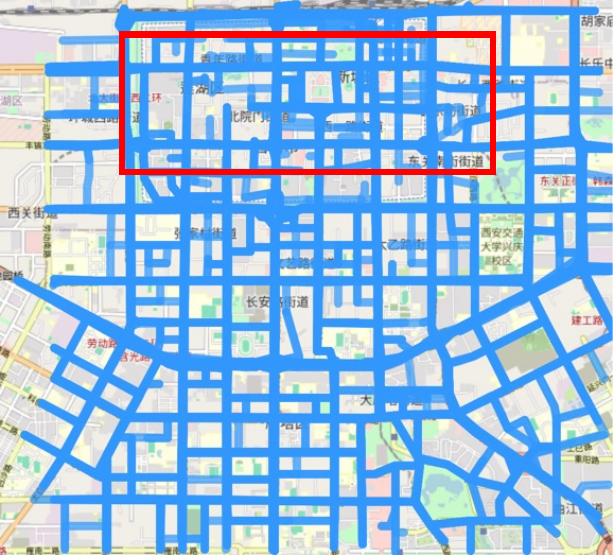}
    \label{subfig:xa_gen_lpd}
    % \Description{Map with routes.}
}
\quad
\subfigure[GPD]{
    \includegraphics[width=0.24\textwidth]{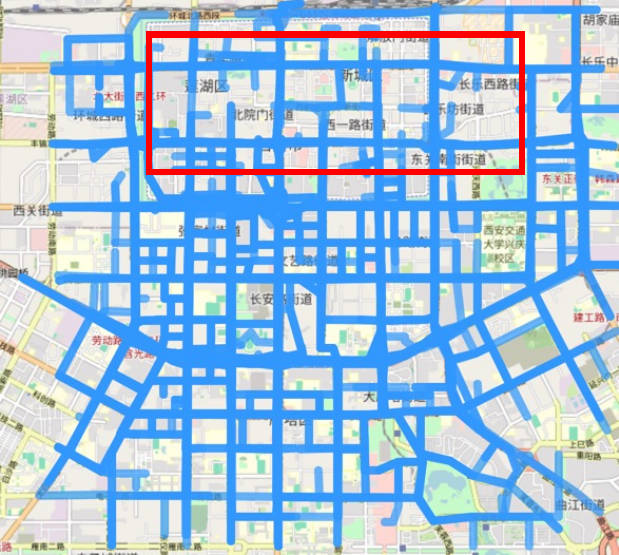}
    \label{subfig:xa_gen_gpd}
    % \Description{Map with routes.}
}
\caption{City B Generated Path Comparison.}
\label{fig:xa_gen}
\end{figure*}

\subsection{Effectiveness Evaluation}
Our \algoname not only achieves great efficiency, but also achieves state-of-the art effectiveness performances.

\begin{table}[htb]
\caption{Unconditional Generation Performance.}
\centering
\begin{tabular}{c|cc|cc}
\multirow{2}{*}{Methods} & \multicolumn{2}{c}{Metrics (Dataset A)} & \multicolumn{2}{c}{Metrics (Dataset B)}  \\
& NLL ($\downarrow$) & NLL-BR ($\uparrow$) & NLL ($\downarrow$) & NLL-BR ($\uparrow$)\\
\hline
MTNet  & $70.59$ &  ---  & $84.59$  & ---   \\
CSSRNN & $28.60$ & $59.3\%$ & $27.26$ & $79.1\%$ \\
GPD    & $27.47$ & $60.9\%$ & $26.38$ & $78.0\%$ \\
\algoname (Ours) & $\mathbf{20.73}$ & $\mathbf{69.0\%}$ & $\mathbf{17.42}$ & $\mathbf{85.8\%}$ \\
% \algoname-DDIM-10 (Ours) & $19.40$ & $71.8\%$ & $16.99$ & $86.1\%$ 
\end{tabular}
\label{tab:uncond-nll}
\end{table}

\fakeparagraph{Unconditional Path Generation}
\tabref{tab:uncond-nll} indicates the unconditional performance comparison.
Our \algoname outperforms other baselines including existing state-of-the-art method GPD.
In terms of mean NLL, our method is lower than others, indicating that \algoname has a better distribution modeling than other methods.
Also, taking MTNet as baseline, our method have the highest BR than other methods, which implies that the our \algoname can steadily beat other baselines.
Concretely, we improve the NLL by $71\%\sim 79\%$ in different datasets.

\figref{fig:cd_gen} and \figref{fig:xa_gen} illustrates generated paths comparisons of two datasets in different cities, respectively.
It can be shown that our method generates paths with more details, as indicated by the red box.
Firstly, our method can capture the path distribution globally.
Similar to real data, some main roads as are covered by more generated paths, denoted as bolder lines.
Secondly, it can be shown that, compared with existing state-of-the-art method GPD, our \algoname has stronger ability to generate paths passing small road segments.
As shown in the red boxes, paths generated by our methods shows more details than GPD.

\begin{figure}[htb]
\centering
\subfigure[Real]{
    \includegraphics[width=0.2\textwidth]{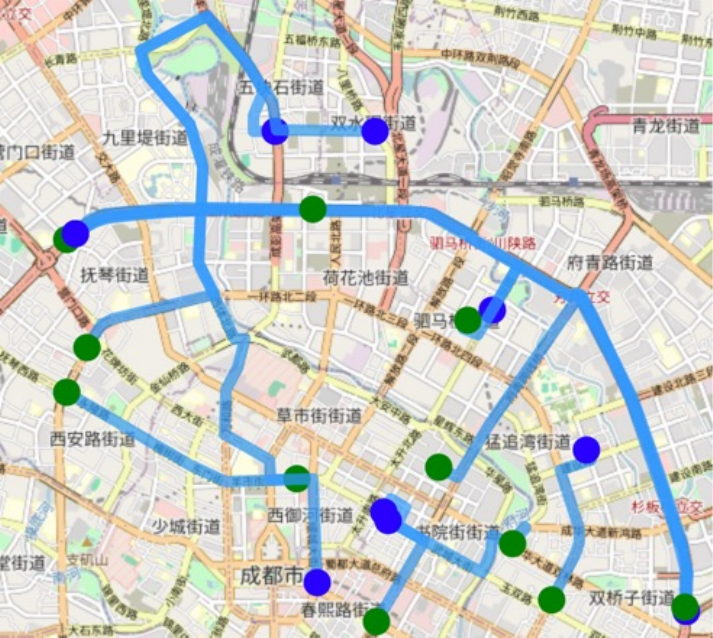}
    \label{subfig:cd_plan_real}
    % \Description{Map with routes.}
}
\subfigure[LPD (Ours)]{
    \includegraphics[width=0.2\textwidth]{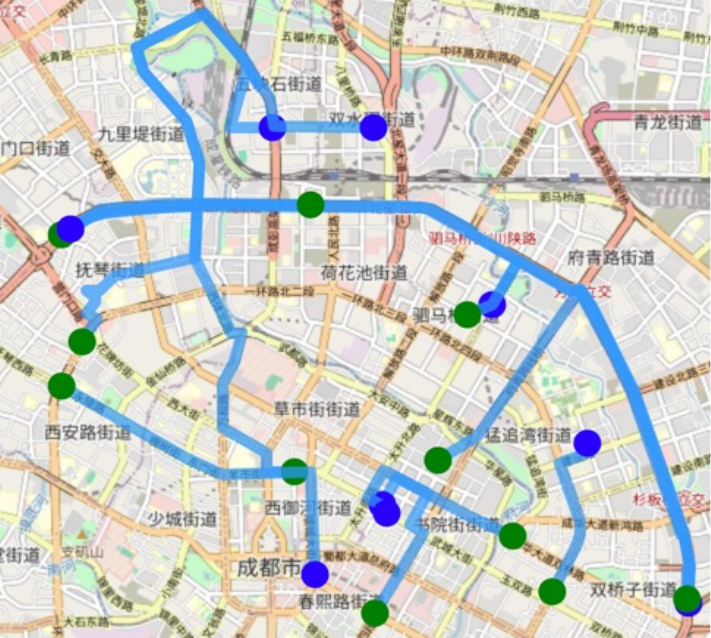}
    \label{subfig:cd_plan_lpd}
    % \Description{Map with routes.}
}
\caption{City A Planned Path Comparison.}
\label{fig:cd_plan}
\end{figure}

\begin{figure}[htb]
\centering
\subfigure[Real]{
    \includegraphics[width=0.2\textwidth]{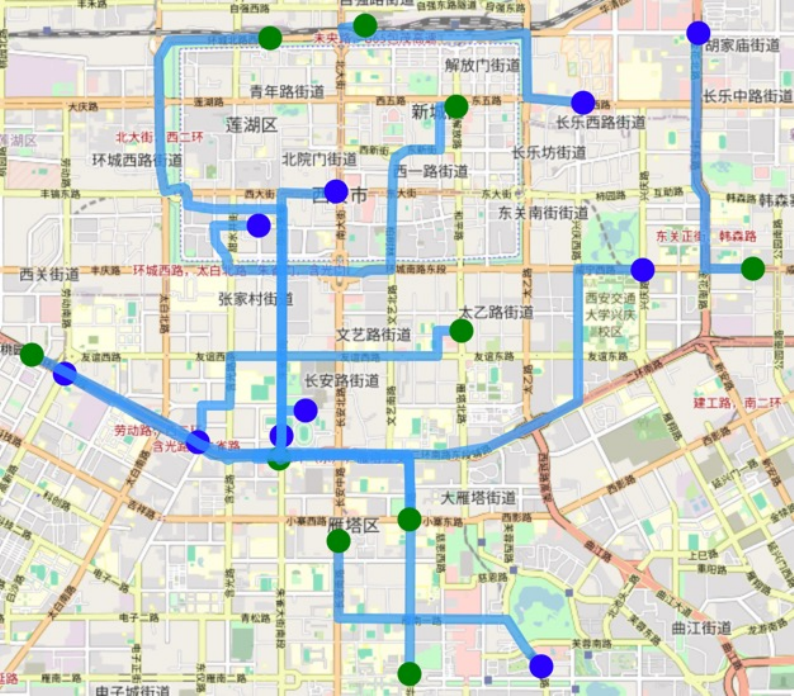}
    \label{subfig:xa_plan_real}
    % \Description{Map with routes.}
}
\subfigure[LPD (Ours)]{
    \includegraphics[width=0.2\textwidth]{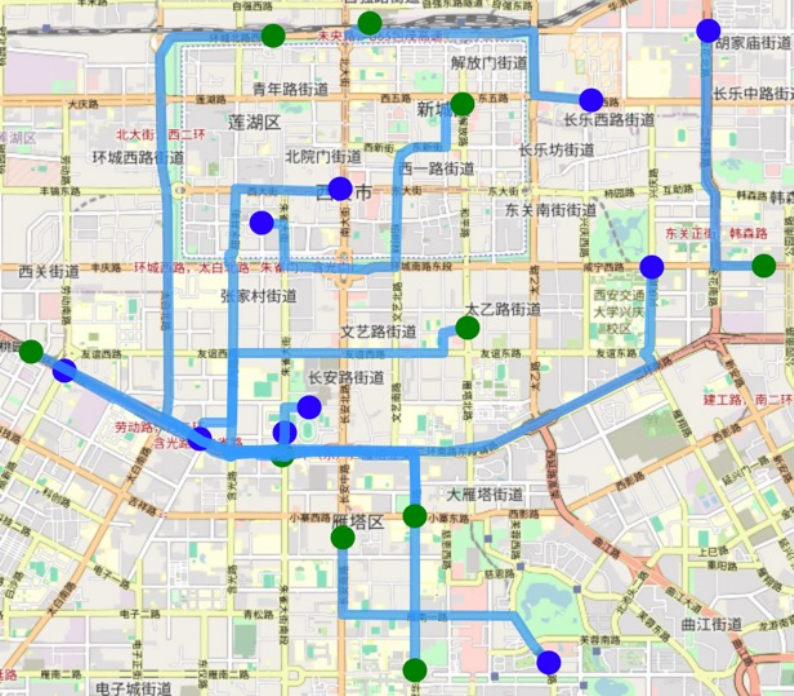}
    \label{subfig:xa_plan_lpd}
    % \Description{Map with routes.}
}
\caption{City B Planned Path Comparison.}
\label{fig:xa_plan}
\end{figure}

\fakeparagraph{Conditional Path Generation}
Since not all methods are based on generative models, some methods do not output probabilities at all.
To make comparison, we measure path similarities between output and ground truth paths.
We use two evaluation metrics, one is longest common sub-sequence (LCS), which view the path as vertex sequence and measure the similarity in terms of vertices.
This metric stress more on the path similarity with regard to graph structure.
Another metric is edit distance with real cost.
It represents the least cost to convert one path to another.
The cost is the distance between two vertices, hence the metric values more on the spatial features.

As for the performance stability, we also adopt beat ratio (BR) for the same reason.
The planning metrics comparison are illustrated in \tabref{tab:cond-nll}, the upper and bottom part indicate values of dataset A and B, respectively.
Overall, our \algoname performs better than other methods in most cases.
In dataset A, our method has the best LCS and LCS-BR, and it achieves comparable performances than previous state-of-the-art method GPD.
In dataset B, our method steadily beats other baselines in both EDR and LCS together with the BR.

\figref{fig:cd_plan} and \figref{fig:xa_plan} provide an illustration for the planned paths.
It can be shown that our method outputs paths that are very similar to ground truth data.

\begin{table*}[htb]
\caption{Conditional Generation Performance.}
\centering
\begin{tabular}{c|cccc}
\toprule
Methods & EDR ($\downarrow$) & EDR-BR ($\uparrow$) & LCS ($\uparrow$) & LCS-BR ($\uparrow$) \\
\midrule
NMLR  & $26.797$ &  ---   & $9.729$ &  --- \\
KS  & $27.637$ & $66.5\%$ & $11.501$ & $62.3\%$ \\
GPD  & $\mathbf{20.306}$ & $67.1\%$ & $14.083$ & $70.1\%$ \\
\algoname(Ours) & $20.379$ & $\mathbf{70.8\%}$ & $\mathbf{15.936}$ & $\mathbf{74.3\%}$ \\
\hline
NMLR  & $31.511$ & --- & $9.712$ & --- \\
KS & $22.956$ & $74.3\%$ & $12.207$ &  $63.4\%$ \\
GPD & $23.049$ & $70.4\%$ & $15.468$ &  $69.5\%$ \\
\algoname(Ours) & $\mathbf{21.730}$ & $\mathbf{77.4\%}$ & $\mathbf{16.427}$ & $\mathbf{71.8\%}$ \\
\bottomrule
\end{tabular}
\label{tab:cond-nll}
\end{table*}

\section{Related Work}
\label{sec:related}

\fakeparagraph{Traditional Path Data Generation}
Path generation is a fundamental building block for many spatiotemporal applications such as path planning \cite{NeurIPS23Xiao}, traffic prediction \cite{VLDB23Tian}, transportation simulation \cite{VLDB22Wang}, and others. 
Leveraging pattern mining from path data can significantly enhance modeling and prediction for these tasks.
Traditional methods often treat paths as sequences of vertices with a Markov property. 
However, a basic Markov chain model \cite{SuttonB98} can suffer from sparsity issues as the order of the Markov model increases. 
To address this, \cite{Ubicomp14Baratchi} proposed using hierarchical hidden states to mitigate the problem. 
Additionally, enhancements like adding contextual information (e.g., weather) have been used to improve performance \cite{TITS18Yin}.
Nevertheless, the strong assumption of the Markov property restricts the flexibility and usability of these traditional methods. 
Recent studies have shifted towards sequence-to-sequence models for better utility. 
Some studies use recurrent neural networks (RNNs) \cite{IJCAI17Wu}, while others employ transformer structures \cite{CIKM22Liang}. 
These neural network-based approaches help overcome the limitations of the Markov property.
To make better use of the graph structure for path generation, some works have explored hierarchical generation strategies \cite{KDD21Fu}, while others employ relay vertex prediction to ease the complexity of generating long paths \cite{VLDB22Wang}.

Recently, with the growing success of diffusion models in image and video generation \cite{ICCV23Peebles, CoRR24Esser}, diffusion-based path generation methods have emerged, as we will introduce next.

\fakeparagraph{Diffusion Models based Path Data Generation}
The diffusion process for generative modeling was first introduced by \cite{ICML15Dickstein}, which creatively employed a series of denoising autoencoders for generative tasks. 
Since then, many works have contributed to improving diffusion models, focusing on various aspects such as modeling the diffusion process \cite{NeurIPS20Ho, ICLR23Liu}, accelerating the sampling process \cite{ICLR21Song}, and enabling conditional generation \cite{CoRR24Ho}. 
These advances have largely targeted perceptual data generation, including images, videos, and audio, achieving significant success.

A major drawback of diffusion models is the high computational cost associated with the large number of timesteps required for the reverse generation process. 
This typically involves simulating ordinary or stochastic differential equations. 
To address this, \cite{ICLR21Song} broke the Markov property by directly modeling the marginal distribution, while some recent studies have moved away from score matching to flow matching \cite{ICLR23Lipman}, with rectified flow \cite{ICLR23Liu} aiming to generate nearly straight diffusion trajectories for one-step generation.

Despite these advancements, much of the focus has been on numerical values, with relatively little attention on categorical values. 
As a result, diffusion models designed for categorical values have lagged behind in adopting the latest innovations.
\cite{NeurIPS21Austin} proposed a general framework for categorical diffusion, while \cite{ICLR24Shi} was the first to design categorical diffusion models in graph space. 
However, these methods are tightly coupled with graph structures, leading to inflexibility and restricting their ability to incorporate the latest advances seen in numerical diffusion models.

% \TODO{refs} propose to adopt autoencoders and conduct diffusion process in latent space.
% Thanks to the autoencoders, the image data can be confined in a variance-restricted, low dimension space.
% This inspired us to devise a latent diffusion process for path data in graph space. 

\section{Conclusion}
\label{sec:conclusion}
In this paper, we propose Latent-space Path Diffusion (\algoname), a highly efficient diffusion model for path generation. 
We identify the primary cause of the low efficiency in the current state-of-the-art method: its diffusion process is explicitly designed to operate in graph space. 
To address this, we develop a latent diffusion model, becoming the first to apply latent diffusion to the path generation problem.
Our experiments with real-world data demonstrate that our approach reduces time and memory costs by as much as 82.8\% and 83.1\%, respectively, while achieving state-of-the-art performance in most scenarios. 
This result confirms the effectiveness and efficiency of using latent diffusion for path generation.
Given the similarities in latent space across various types of data, we believe our work can serve as a bridge, connecting path generation with the extensive advancements made in latent diffusion models for perceptual data generation. 
This connection could open up new avenues for research and application, benefiting from the broader experiences in the field of diffusion-based generative modeling.
We point out that the limitation of our work is that we trained models for different road networks, but we think this limitation can be solved in latent space and view it as a future work.

\balance
\bibliographystyle{ACM-Reference-Format}
\bibliography{main}


\begin{thebibliography}{33}

%%% the Web address.

\ifx \showCODEN    \undefined \def \showCODEN     #1{\unskip}     \fi
\ifx \showDOI      \undefined \def \showDOI       #1{#1}\fi
\ifx \showISBNx    \undefined \def \showISBNx     #1{\unskip}     \fi
\ifx \showISBNxiii \undefined \def \showISBNxiii  #1{\unskip}     \fi
\ifx \showISSN     \undefined \def \showISSN      #1{\unskip}     \fi
\ifx \showLCCN     \undefined \def \showLCCN      #1{\unskip}     \fi
\ifx \shownote     \undefined \def \shownote      #1{#1}          \fi
\ifx \showarticletitle \undefined \def \showarticletitle #1{#1}   \fi
\ifx \showURL      \undefined \def \showURL       {\relax}        \fi
% invisible to TeX
\providecommand\bibfield[2]{#2}
\providecommand\bibinfo[2]{#2}
\providecommand\natexlab[1]{#1}
\providecommand\showeprint[2][]{arXiv:#2}

\bibitem[Austin et~al\mbox{.}(2021)]%
        {NeurIPS21Austin}
\bibfield{author}{\bibinfo{person}{Jacob Austin}, \bibinfo{person}{Daniel~D. Johnson}, \bibinfo{person}{Jonathan Ho}, \bibinfo{person}{Daniel Tarlow}, {and} \bibinfo{person}{Rianne van~den Berg}.} \bibinfo{year}{2021}\natexlab{}.
\newblock \showarticletitle{Structured Denoising Diffusion Models in Discrete State-Spaces}. In \bibinfo{booktitle}{\emph{Advances in Neural Information Processing Systems 34: Annual Conference on Neural Information Processing Systems 2021, NeurIPS 2021, December 6-14, 2021, virtual}}. \bibinfo{publisher}{neurips.cc}, \bibinfo{address}{Virtual}, \bibinfo{pages}{17981--17993}.
\newblock


\bibitem[Baratchi et~al\mbox{.}(2014)]%
        {Ubicomp14Baratchi}
\bibfield{author}{\bibinfo{person}{Mitra Baratchi}, \bibinfo{person}{Nirvana Meratnia}, \bibinfo{person}{Paul J.~M. Havinga}, \bibinfo{person}{Andrew~K. Skidmore}, {and} \bibinfo{person}{Bert A.~G. Toxopeus}.} \bibinfo{year}{2014}\natexlab{}.
\newblock \showarticletitle{A hierarchical hidden semi-Markov model for modeling mobility data}. In \bibinfo{booktitle}{\emph{The 2014 {ACM} Conference on Ubiquitous Computing, UbiComp '14, Seattle, WA, USA, September 13-17, 2014}}, \bibfield{editor}{\bibinfo{person}{A.~J. Brush}, \bibinfo{person}{Adrian Friday}, \bibinfo{person}{Julie~A. Kientz}, \bibinfo{person}{James Scott}, {and} \bibinfo{person}{Junehwa Song}} (Eds.). \bibinfo{publisher}{{ACM}}, \bibinfo{address}{USA}, \bibinfo{pages}{401--412}.
\newblock


\bibitem[Brown et~al\mbox{.}(2020)]%
        {NeurIPS20Brown}
\bibfield{author}{\bibinfo{person}{Tom~B. Brown}, \bibinfo{person}{Benjamin Mann}, \bibinfo{person}{Nick Ryder}, \bibinfo{person}{Melanie Subbiah}, \bibinfo{person}{Jared Kaplan}, \bibinfo{person}{Prafulla Dhariwal}, \bibinfo{person}{Arvind Neelakantan}, \bibinfo{person}{Pranav Shyam}, \bibinfo{person}{Girish Sastry}, \bibinfo{person}{Amanda Askell}, \bibinfo{person}{Sandhini Agarwal}, \bibinfo{person}{Ariel Herbert{-}Voss}, \bibinfo{person}{Gretchen Krueger}, \bibinfo{person}{Tom Henighan}, \bibinfo{person}{Rewon Child}, \bibinfo{person}{Aditya Ramesh}, \bibinfo{person}{Daniel~M. Ziegler}, \bibinfo{person}{Jeffrey Wu}, \bibinfo{person}{Clemens Winter}, \bibinfo{person}{Christopher Hesse}, \bibinfo{person}{Mark Chen}, \bibinfo{person}{Eric Sigler}, \bibinfo{person}{Mateusz Litwin}, \bibinfo{person}{Scott Gray}, \bibinfo{person}{Benjamin Chess}, \bibinfo{person}{Jack Clark}, \bibinfo{person}{Christopher Berner}, \bibinfo{person}{Sam McCandlish}, \bibinfo{person}{Alec Radford}, \bibinfo{person}{Ilya Sutskever},
  {and} \bibinfo{person}{Dario Amodei}.} \bibinfo{year}{2020}\natexlab{}.
\newblock \showarticletitle{Language Models are Few-Shot Learners}. In \bibinfo{booktitle}{\emph{Advances in Neural Information Processing Systems 33: Annual Conference on Neural Information Processing Systems 2020, NeurIPS 2020, December 6-12, 2020, virtual}}. \bibinfo{publisher}{neurips.cc}, \bibinfo{address}{Virtual}.
\newblock


\bibitem[Devlin et~al\mbox{.}(2019)]%
        {NAACL19Devlin}
\bibfield{author}{\bibinfo{person}{Jacob Devlin}, \bibinfo{person}{Ming{-}Wei Chang}, \bibinfo{person}{Kenton Lee}, {and} \bibinfo{person}{Kristina Toutanova}.} \bibinfo{year}{2019}\natexlab{}.
\newblock \showarticletitle{{BERT:} Pre-training of Deep Bidirectional Transformers for Language Understanding}. In \bibinfo{booktitle}{\emph{Proceedings of the 2019 Conference of the North American Chapter of the Association for Computational Linguistics: Human Language Technologies, {NAACL-HLT} 2019, Minneapolis, MN, USA, June 2-7, 2019, Volume 1 (Long and Short Papers)}}. \bibinfo{publisher}{Association for Computational Linguistics}, \bibinfo{address}{USA}, \bibinfo{pages}{4171--4186}.
\newblock


\bibitem[Dosovitskiy et~al\mbox{.}(2021)]%
        {ICLR21Dosovitskiy}
\bibfield{author}{\bibinfo{person}{Alexey Dosovitskiy}, \bibinfo{person}{Lucas Beyer}, \bibinfo{person}{Alexander Kolesnikov}, \bibinfo{person}{Dirk Weissenborn}, \bibinfo{person}{Xiaohua Zhai}, \bibinfo{person}{Thomas Unterthiner}, \bibinfo{person}{Mostafa Dehghani}, \bibinfo{person}{Matthias Minderer}, \bibinfo{person}{Georg Heigold}, \bibinfo{person}{Sylvain Gelly}, \bibinfo{person}{Jakob Uszkoreit}, {and} \bibinfo{person}{Neil Houlsby}.} \bibinfo{year}{2021}\natexlab{}.
\newblock \showarticletitle{An Image is Worth 16x16 Words: Transformers for Image Recognition at Scale}. In \bibinfo{booktitle}{\emph{9th International Conference on Learning Representations, {ICLR} 2021, Virtual Event, Austria, May 3-7, 2021}}. \bibinfo{publisher}{OpenReview.net}, \bibinfo{address}{Austria}.
\newblock


\bibitem[Esser et~al\mbox{.}(2024)]%
        {CoRR24Esser}
\bibfield{author}{\bibinfo{person}{Patrick Esser}, \bibinfo{person}{Sumith Kulal}, \bibinfo{person}{Andreas Blattmann}, \bibinfo{person}{Rahim Entezari}, \bibinfo{person}{Jonas M{\"{u}}ller}, \bibinfo{person}{Harry Saini}, \bibinfo{person}{Yam Levi}, \bibinfo{person}{Dominik Lorenz}, \bibinfo{person}{Axel Sauer}, \bibinfo{person}{Frederic Boesel}, \bibinfo{person}{Dustin Podell}, \bibinfo{person}{Tim Dockhorn}, \bibinfo{person}{Zion English}, \bibinfo{person}{Kyle Lacey}, \bibinfo{person}{Alex Goodwin}, \bibinfo{person}{Yannik Marek}, {and} \bibinfo{person}{Robin Rombach}.} \bibinfo{year}{2024}\natexlab{}.
\newblock \showarticletitle{Scaling Rectified Flow Transformers for High-Resolution Image Synthesis}.
\newblock \bibinfo{journal}{\emph{CoRR}}  \bibinfo{volume}{abs/2403.03206} (\bibinfo{year}{2024}).
\newblock


\bibitem[Fu and Lee(2021)]%
        {KDD21Fu}
\bibfield{author}{\bibinfo{person}{Tao{-}Yang Fu} {and} \bibinfo{person}{Wang{-}Chien Lee}.} \bibinfo{year}{2021}\natexlab{}.
\newblock \showarticletitle{ProgRPGAN: Progressive {GAN} for Route Planning}. In \bibinfo{booktitle}{\emph{{KDD} '21: The 27th {ACM} {SIGKDD} Conference on Knowledge Discovery and Data Mining, Virtual Event, Singapore, August 14-18, 2021}}. \bibinfo{publisher}{{ACM}}, \bibinfo{address}{Singapore}, \bibinfo{pages}{393--403}.
\newblock


\bibitem[Ho et~al\mbox{.}(2020)]%
        {NeurIPS20Ho}
\bibfield{author}{\bibinfo{person}{Jonathan Ho}, \bibinfo{person}{Ajay Jain}, {and} \bibinfo{person}{Pieter Abbeel}.} \bibinfo{year}{2020}\natexlab{}.
\newblock \showarticletitle{Denoising Diffusion Probabilistic Models}. In \bibinfo{booktitle}{\emph{Advances in Neural Information Processing Systems 33: Annual Conference on Neural Information Processing Systems 2020, NeurIPS 2020, December 12, 2020, virtual}}. \bibinfo{publisher}{neurips.cc}, \bibinfo{address}{Virtual}.
\newblock


\bibitem[Ho and Salimans(2022a)]%
        {CoRR22Ho}
\bibfield{author}{\bibinfo{person}{Jonathan Ho} {and} \bibinfo{person}{Tim Salimans}.} \bibinfo{year}{2022}\natexlab{a}.
\newblock \showarticletitle{Classifier-Free Diffusion Guidance}.
\newblock \bibinfo{journal}{\emph{CoRR}}  \bibinfo{volume}{abs/2207.12598} (\bibinfo{year}{2022}).
\newblock


\bibitem[Ho and Salimans(2022b)]%
        {CoRR24Ho}
\bibfield{author}{\bibinfo{person}{Jonathan Ho} {and} \bibinfo{person}{Tim Salimans}.} \bibinfo{year}{2022}\natexlab{b}.
\newblock \showarticletitle{Classifier-Free Diffusion Guidance}.
\newblock \bibinfo{journal}{\emph{CoRR}}  \bibinfo{volume}{abs/2207.12598} (\bibinfo{year}{2022}).
\newblock
\urldef\tempurl%
\url{https://doi.org/10.48550/ARXIV.2207.12598}
\showDOI{\tempurl}


\bibitem[Hong et~al\mbox{.}(2020)]%
        {KDD20Hong}
\bibfield{author}{\bibinfo{person}{Huiting Hong}, \bibinfo{person}{Yucheng Lin}, \bibinfo{person}{Xiaoqing Yang}, \bibinfo{person}{Zang Li}, \bibinfo{person}{Kun Fu}, \bibinfo{person}{Zheng Wang}, \bibinfo{person}{Xiaohu Qie}, {and} \bibinfo{person}{Jieping Ye}.} \bibinfo{year}{2020}\natexlab{}.
\newblock \showarticletitle{HetETA: Heterogeneous Information Network Embedding for Estimating Time of Arrival}. In \bibinfo{booktitle}{\emph{{KDD} '20: The 26th {ACM} {SIGKDD} Conference on Knowledge Discovery and Data Mining, Virtual Event, CA, USA, August 23-27, 2020}}. \bibinfo{publisher}{{ACM}}, \bibinfo{address}{USA}, \bibinfo{pages}{2444--2454}.
\newblock


\bibitem[Jain et~al\mbox{.}(2021)]%
        {NeurIPS21Jain}
\bibfield{author}{\bibinfo{person}{Jayant Jain}, \bibinfo{person}{Vrittika Bagadia}, \bibinfo{person}{Sahil Manchanda}, {and} \bibinfo{person}{Sayan Ranu}.} \bibinfo{year}{2021}\natexlab{}.
\newblock \showarticletitle{NeuroMLR: Robust {\&} Reliable Route Recommendation on Road Networks}. In \bibinfo{booktitle}{\emph{Advances in Neural Information Processing Systems 34: Annual Conference on Neural Information Processing Systems 2021, NeurIPS 2021, December 6-14, 2021, virtual}}. \bibinfo{publisher}{neurips.cc}, \bibinfo{address}{Virtual}, \bibinfo{pages}{22070--22082}.
\newblock


\bibitem[Janner et~al\mbox{.}(2022)]%
        {ICML22Janner}
\bibfield{author}{\bibinfo{person}{Michael Janner}, \bibinfo{person}{Yilun Du}, \bibinfo{person}{Joshua~B. Tenenbaum}, {and} \bibinfo{person}{Sergey Levine}.} \bibinfo{year}{2022}\natexlab{}.
\newblock \showarticletitle{Planning with Diffusion for Flexible Behavior Synthesis}. In \bibinfo{booktitle}{\emph{International Conference on Machine Learning, {ICML} 2022, 17-23 July 2022, Baltimore, Maryland, {USA}}} \emph{(\bibinfo{series}{Proceedings of Machine Learning Research}, Vol.~\bibinfo{volume}{162})}. \bibinfo{publisher}{{PMLR}}, \bibinfo{address}{USA}, \bibinfo{pages}{9902--9915}.
\newblock


\bibitem[Jiang et~al\mbox{.}(2023)]%
        {CVPR23Jiang}
\bibfield{author}{\bibinfo{person}{Chiyu~Max Jiang}, \bibinfo{person}{Andre Cornman}, \bibinfo{person}{Cheolho Park}, \bibinfo{person}{Benjamin Sapp}, \bibinfo{person}{Yin Zhou}, {and} \bibinfo{person}{Dragomir Anguelov}.} \bibinfo{year}{2023}\natexlab{}.
\newblock \showarticletitle{MotionDiffuser: Controllable Multi-Agent Motion Prediction Using Diffusion}. In \bibinfo{booktitle}{\emph{{IEEE/CVF} Conference on Computer Vision and Pattern Recognition, {CVPR} 2023, Vancouver, BC, Canada, June 17-24, 2023}}. \bibinfo{publisher}{{IEEE}}, \bibinfo{address}{Canada}, \bibinfo{pages}{9644--9653}.
\newblock


\bibitem[Kingma and Welling(2014)]%
        {ICLR14Kingma}
\bibfield{author}{\bibinfo{person}{Diederik~P. Kingma} {and} \bibinfo{person}{Max Welling}.} \bibinfo{year}{2014}\natexlab{}.
\newblock \showarticletitle{Auto-Encoding Variational Bayes}. In \bibinfo{booktitle}{\emph{2nd International Conference on Learning Representations, {ICLR} 2014, Banff, AB, Canada, April 14-16, 2014, Conference Track Proceedings}}. \bibinfo{publisher}{OpenReview.net}, \bibinfo{address}{Canada}.
\newblock


\bibitem[Liang et~al\mbox{.}(2022)]%
        {CIKM22Liang}
\bibfield{author}{\bibinfo{person}{Yuxuan Liang}, \bibinfo{person}{Kun Ouyang}, \bibinfo{person}{Yiwei Wang}, \bibinfo{person}{Xu Liu}, \bibinfo{person}{Hongyang Chen}, \bibinfo{person}{Junbo Zhang}, \bibinfo{person}{Yu Zheng}, {and} \bibinfo{person}{Roger Zimmermann}.} \bibinfo{year}{2022}\natexlab{}.
\newblock \showarticletitle{TrajFormer: Efficient Trajectory Classification with Transformers}. In \bibinfo{booktitle}{\emph{Proceedings of the 31st {ACM} International Conference on Information {\&} Knowledge Management, Atlanta, GA, USA, October 17-21, 2022}}, \bibfield{editor}{\bibinfo{person}{Mohammad~Al Hasan} {and} \bibinfo{person}{Li~Xiong}} (Eds.). \bibinfo{publisher}{{ACM}}, \bibinfo{address}{USA}, \bibinfo{pages}{1229--1237}.
\newblock


\bibitem[Lipman et~al\mbox{.}(2023)]%
        {ICLR23Lipman}
\bibfield{author}{\bibinfo{person}{Yaron Lipman}, \bibinfo{person}{Ricky T.~Q. Chen}, \bibinfo{person}{Heli Ben{-}Hamu}, \bibinfo{person}{Maximilian Nickel}, {and} \bibinfo{person}{Matthew Le}.} \bibinfo{year}{2023}\natexlab{}.
\newblock \showarticletitle{Flow Matching for Generative Modeling}. In \bibinfo{booktitle}{\emph{The Eleventh International Conference on Learning Representations, {ICLR} 2023, Kigali, Rwanda, May 1-5, 2023}}. \bibinfo{publisher}{OpenReview.net}, \bibinfo{address}{Rwanda}.
\newblock


\bibitem[Liu et~al\mbox{.}(2023)]%
        {ICLR23Liu}
\bibfield{author}{\bibinfo{person}{Xingchao Liu}, \bibinfo{person}{Chengyue Gong}, {and} \bibinfo{person}{Qiang Liu}.} \bibinfo{year}{2023}\natexlab{}.
\newblock \showarticletitle{Flow Straight and Fast: Learning to Generate and Transfer Data with Rectified Flow}. In \bibinfo{booktitle}{\emph{The Eleventh International Conference on Learning Representations, {ICLR} 2023, Kigali, Rwanda, May 1-5, 2023}}. \bibinfo{publisher}{OpenReview.net}, \bibinfo{address}{Rwanda}.
\newblock


\bibitem[Meert and Verbeke(2018)]%
        {ECDA18Meert}
\bibfield{author}{\bibinfo{person}{Wannes Meert} {and} \bibinfo{person}{Mathias Verbeke}.} \bibinfo{year}{2018}\natexlab{}.
\newblock \showarticletitle{HMM with non-emitting states for Map Matching}. In \bibinfo{booktitle}{\emph{European Conference on Data Analysis (ECDA)}}.
\newblock


\bibitem[Peebles and Xie(2023)]%
        {ICCV23Peebles}
\bibfield{author}{\bibinfo{person}{William Peebles} {and} \bibinfo{person}{Saining Xie}.} \bibinfo{year}{2023}\natexlab{}.
\newblock \showarticletitle{Scalable Diffusion Models with Transformers}. In \bibinfo{booktitle}{\emph{{IEEE/CVF} International Conference on Computer Vision, {ICCV} 2023, Paris, France, October 1-6, 2023}}. \bibinfo{publisher}{{IEEE}}, \bibinfo{address}{France}, \bibinfo{pages}{4172--4182}.
\newblock


\bibitem[Rombach et~al\mbox{.}(2022)]%
        {CVPR22Rombach}
\bibfield{author}{\bibinfo{person}{Robin Rombach}, \bibinfo{person}{Andreas Blattmann}, \bibinfo{person}{Dominik Lorenz}, \bibinfo{person}{Patrick Esser}, {and} \bibinfo{person}{Bj{\"{o}}rn Ommer}.} \bibinfo{year}{2022}\natexlab{}.
\newblock \showarticletitle{High-Resolution Image Synthesis with Latent Diffusion Models}. In \bibinfo{booktitle}{\emph{{IEEE/CVF} Conference on Computer Vision and Pattern Recognition, {CVPR} 2022, New Orleans, LA, USA, June 18-24, 2022}}. \bibinfo{publisher}{{IEEE}}, \bibinfo{address}{USA}, \bibinfo{pages}{10674--10685}.
\newblock


\bibitem[Shi et~al\mbox{.}(2024)]%
        {ICLR24Shi}
\bibfield{author}{\bibinfo{person}{Dingyuan Shi}, \bibinfo{person}{Yongxin Tong}, \bibinfo{person}{Zimu Zhou}, \bibinfo{person}{Ke Xu}, \bibinfo{person}{Zheng Wang}, {and} \bibinfo{person}{Jieping Ye}.} \bibinfo{year}{2024}\natexlab{}.
\newblock \showarticletitle{Graph-constrained diffusion for End-to-End Path Planning}. In \bibinfo{booktitle}{\emph{The Twelfth International Conference on Learning Representations, {ICLR} 2024, Vienna, Austria, May 7-11, 2024}}. \bibinfo{publisher}{OpenReview.net}, \bibinfo{address}{Austria}.
\newblock


\bibitem[Sohl{-}Dickstein et~al\mbox{.}(2015)]%
        {ICML15Dickstein}
\bibfield{author}{\bibinfo{person}{Jascha Sohl{-}Dickstein}, \bibinfo{person}{Eric~A. Weiss}, \bibinfo{person}{Niru Maheswaranathan}, {and} \bibinfo{person}{Surya Ganguli}.} \bibinfo{year}{2015}\natexlab{}.
\newblock \showarticletitle{Deep Unsupervised Learning using Nonequilibrium Thermodynamics}. In \bibinfo{booktitle}{\emph{Proceedings of the 32nd International Conference on Machine Learning, {ICML} 2015, Lille, France, 6-11 July 2015}} \emph{(\bibinfo{series}{{JMLR} Workshop and Conference Proceedings}, Vol.~\bibinfo{volume}{37})}. \bibinfo{publisher}{JMLR.org}, \bibinfo{address}{France}, \bibinfo{pages}{2256--2265}.
\newblock


\bibitem[Song et~al\mbox{.}(2021)]%
        {ICLR21Song}
\bibfield{author}{\bibinfo{person}{Jiaming Song}, \bibinfo{person}{Chenlin Meng}, {and} \bibinfo{person}{Stefano Ermon}.} \bibinfo{year}{2021}\natexlab{}.
\newblock \showarticletitle{Denoising Diffusion Implicit Models}. In \bibinfo{booktitle}{\emph{9th International Conference on Learning Representations, {ICLR} 2021, Virtual Event, Austria, May 3-7, 2021}}. \bibinfo{publisher}{OpenReview.net}, \bibinfo{address}{Austria}.
\newblock


\bibitem[Sutton and Barto(1998)]%
        {SuttonB98}
\bibfield{author}{\bibinfo{person}{Richard~S. Sutton} {and} \bibinfo{person}{Andrew~G. Barto}.} \bibinfo{year}{1998}\natexlab{}.
\newblock \bibinfo{booktitle}{\emph{Reinforcement learning - an introduction}}.
\newblock \bibinfo{publisher}{{MIT} Press}, \bibinfo{address}{USA}.
\newblock


\bibitem[Tian et~al\mbox{.}(2023)]%
        {VLDB23Tian}
\bibfield{author}{\bibinfo{person}{Wei Tian}, \bibinfo{person}{Jieming Shi}, \bibinfo{person}{Siqiang Luo}, \bibinfo{person}{Hui Li}, \bibinfo{person}{Xike Xie}, {and} \bibinfo{person}{Yuanhang Zou}.} \bibinfo{year}{2023}\natexlab{}.
\newblock \showarticletitle{Effective and Efficient Route Planning Using Historical Trajectories on Road Networks}.
\newblock \bibinfo{journal}{\emph{Proceedings of the VLDB Endowment}} \bibinfo{volume}{16}, \bibinfo{number}{10} (\bibinfo{year}{2023}), \bibinfo{pages}{2512--2524}.
\newblock


\bibitem[Vaswani et~al\mbox{.}(2017)]%
        {NeurIPS17Vaswani}
\bibfield{author}{\bibinfo{person}{Ashish Vaswani}, \bibinfo{person}{Noam Shazeer}, \bibinfo{person}{Niki Parmar}, \bibinfo{person}{Jakob Uszkoreit}, \bibinfo{person}{Llion Jones}, \bibinfo{person}{Aidan~N. Gomez}, \bibinfo{person}{Lukasz Kaiser}, {and} \bibinfo{person}{Illia Polosukhin}.} \bibinfo{year}{2017}\natexlab{}.
\newblock \showarticletitle{Attention is All you Need}. In \bibinfo{booktitle}{\emph{Advances in Neural Information Processing Systems 30: Annual Conference on Neural Information Processing Systems 2017, December 4-9, 2017, Long Beach, CA, {USA}}}. \bibinfo{publisher}{neurips.cc}, \bibinfo{address}{USA}, \bibinfo{pages}{5998--6008}.
\newblock


\bibitem[Wang et~al\mbox{.}(2022)]%
        {VLDB22Wang}
\bibfield{author}{\bibinfo{person}{Yong Wang}, \bibinfo{person}{Guoliang Li}, \bibinfo{person}{Kaiyu Li}, {and} \bibinfo{person}{Haitao Yuan}.} \bibinfo{year}{2022}\natexlab{}.
\newblock \showarticletitle{A Deep Generative Model for Trajectory Modeling and Utilization}.
\newblock \bibinfo{journal}{\emph{Proc. {VLDB} Endow.}} \bibinfo{volume}{16}, \bibinfo{number}{4} (\bibinfo{year}{2022}), \bibinfo{pages}{973--985}.
\newblock


\bibitem[Wu et~al\mbox{.}(2017)]%
        {IJCAI17Wu}
\bibfield{author}{\bibinfo{person}{Hao Wu}, \bibinfo{person}{Ziyang Chen}, \bibinfo{person}{Weiwei Sun}, \bibinfo{person}{Baihua Zheng}, {and} \bibinfo{person}{Wei Wang}.} \bibinfo{year}{2017}\natexlab{}.
\newblock \showarticletitle{Modeling Trajectories with Recurrent Neural Networks}. In \bibinfo{booktitle}{\emph{Proceedings of the Twenty-Sixth International Joint Conference on Artificial Intelligence, {IJCAI} 2017, Melbourne, Australia, August 19-25, 2017}}, \bibfield{editor}{\bibinfo{person}{Carles Sierra}} (Ed.). \bibinfo{publisher}{ijcai.org}, \bibinfo{address}{Australia}, \bibinfo{pages}{3083--3090}.
\newblock


\bibitem[Wu et~al\mbox{.}(2018)]%
        {TRPC18Wu}
\bibfield{author}{\bibinfo{person}{Yuankai Wu}, \bibinfo{person}{Huachun Tan}, \bibinfo{person}{Lingqiao Qin}, \bibinfo{person}{Bin Ran}, {and} \bibinfo{person}{Zhuxi Jiang}.} \bibinfo{year}{2018}\natexlab{}.
\newblock \showarticletitle{A hybrid deep learning based traffic flow prediction method and its understanding}.
\newblock \bibinfo{journal}{\emph{Transportation Research Part C: Emerging Technologies}}  \bibinfo{volume}{90} (\bibinfo{year}{2018}), \bibinfo{pages}{166--180}.
\newblock


\bibitem[Yin et~al\mbox{.}(2018)]%
        {TITS18Yin}
\bibfield{author}{\bibinfo{person}{Mogeng Yin}, \bibinfo{person}{Madeleine Sheehan}, \bibinfo{person}{Sidney Feygin}, \bibinfo{person}{Jean{-}Fran{\c{c}}ois Paiement}, {and} \bibinfo{person}{Alexei Pozdnoukhov}.} \bibinfo{year}{2018}\natexlab{}.
\newblock \showarticletitle{A Generative Model of Urban Activities from Cellular Data}.
\newblock \bibinfo{journal}{\emph{{IEEE} Trans. Intell. Transp. Syst.}} \bibinfo{volume}{19}, \bibinfo{number}{6} (\bibinfo{year}{2018}), \bibinfo{pages}{1682--1696}.
\newblock


\bibitem[Zang et~al\mbox{.}(2024)]%
        {NeurIPS23Xiao}
\bibfield{author}{\bibinfo{person}{Xiao Zang}, \bibinfo{person}{Miao Yin}, \bibinfo{person}{Jinqi Xiao}, \bibinfo{person}{Saman Zonouz}, {and} \bibinfo{person}{Bo Yuan}.} \bibinfo{year}{2024}\natexlab{}.
\newblock \showarticletitle{GraphMP: Graph Neural Network-based Motion Planning with Efficient Graph Search}.
\newblock \bibinfo{journal}{\emph{Advances in Neural Information Processing Systems}}  \bibinfo{volume}{36} (\bibinfo{year}{2024}).
\newblock


\bibitem[Zhu et~al\mbox{.}(2023)]%
        {CoRR23Zhu}
\bibfield{author}{\bibinfo{person}{Yuanshao Zhu}, \bibinfo{person}{Yongchao Ye}, \bibinfo{person}{Xiangyu Zhao}, {and} \bibinfo{person}{James J.~Q. Yu}.} \bibinfo{year}{2023}\natexlab{}.
\newblock \showarticletitle{Diffusion Model for {GPS} Trajectory Generation}.
\newblock \bibinfo{journal}{\emph{CoRR}}  \bibinfo{volume}{abs/2304.11582} (\bibinfo{year}{2023}).
\newblock


\end{thebibliography}

\appendix
\clearpage
\section*{Appendix}
\appendix

\section{Proof}
\label{app:proof}
\proposition[]{As $t$ increases, $Q_t$ becomes denser.}

\fakeparagraph{Proof}
$Q_t$ is a matrix exponential, which is calculated as below
\begin{equation}
Q_t=e^{(\mathbf{A}-\mathbf{D})t}= \sum_{k=0}^\infty \frac{(\mathbf{A}-\mathbf{D})^kt^k}{k!}
\end{equation}
We denote $D^{1/2}$ as a diagonal matrix whose element of the $i$-th row and $i$-th column is the square root of degree of vertex $i$, \ie $D^{1/2}_{ii} = \sqrt{d_{i}}$.
Accordingly, $D^{-1/2}$ is the inverse of $D^{1/2}$ and we have $D^{-1/2}_{ii}=\frac{1}{\sqrt{d_i}} \forall i$.
Based on the above auxiliary matrix, we have
\begin{equation}
A-D = D^{-1/2} L_sD^{-1/2}
\end{equation}
where the diagonal elements of matrix $L_s$ are all $-1$ and if vertex $i$ and $j$ are adjacent, $L_s[i,j]=\frac{1}{\sqrt{d_id_j}}$.
we further denote $L_s = L_n - I$, then all diagonal element of $L_n$ is zero, and the $Q_t$ can be rewritten as below
\begin{equation}
\begin{aligned}
Q_t &= \sum_{k=0}^\infty \frac{(D^{-1/2}(L_n-I)D^{-1/2})^k t^k}{k!}  \\
&= \sum_{k=0}^\infty \frac{D^{-k/2}(L_n-I)^kD^{-k/2}t^k}{k!}
\end{aligned}
\end{equation}
Except for $(L_n-I)^k$, other terms are simply diagonal or numbers.
This term can be written as a summation of a series of power of $L_n$
\begin{equation}
(L_n-I)^k = \sum_{i=0}^k C_k^i L_n^i
\end{equation}
Note that the nonzero elements in $L_n$ has the same position as adjacent matrix $A$.
Recall that the nonzero elements of power of adjacent matrix $A$ indicating k-hop connection between two vertices.
By assuming the graph is connected, as the power increases, the k-hop connection increases, and nonzero elements increases, thus, this matrix is not a sparse matrix.

\qed

\proposition[]{$Q_t$ has large ranks.}

\fakeparagraph{Proof}
We denote $H_t = \sum_{k=1}^\infty \frac{(\mathbf{A}-\mathbf{D})^kt^k}{k!}$, then $Q_t = I + H_t$.

Consider a row vector $r_i = [0, 0, ..., 1, ..., 0]$, \ie only the $i$-th element is value 1 otherwise 0.
This row vector represents that the initial heat are all condense at vertex $i$.
Then we multiply it with $Q_t$, which is a simulation of heat conduction after time $t$, so we have
\begin{equation}
r_i Q_t = r_i (I + H_t) = r_i + r_i H_t = [\Delta'_1, ...,1-\Delta,...,\Delta'_{|V|}]
\end{equation}
Since the initial heat is all condensed on vertex $i$, the heat can only flow out from source $i$, so $\Delta > 0$. 
By subtracting $r_i$ at both lhs and rhs, we have
\begin{equation}
    r_i H_t = [\Delta'_1,...,-\Delta, ..., \Delta'_{|V|}]
\end{equation}
Also, based on the property of $Q_t$ from \cite{ICLR24Shi}, $Q_t$ is symmetric and the summation of each row or column equals to 1.
So the summation of all other elements except for $i$-th element equals to $\Delta$ (\ie $\sum_{j\ne i}\Delta'_j=\Delta$).
That means, the $i$-th row of the matrix $H_t$ denotes the specific variation proportion of heat transmitted from source $i$ to others, which is also align with energy conservation law.

Now let us consider the element changing by $t$.
Taking the deriviation of the matrix by $t$, we have
\begin{equation}
\frac{d}{dt} \exp\{(A-D)t\} = (A-D) (I + H_t) \\
= (A-D) + (A-D)H_t
\end{equation}
We only care about the diagonal element changing, the $i$-th element in the $i$-th row will be $-d_i -d_i * \Delta + (\sum_j A_{ij} \Delta'_j) < 0$, so the diagonal elements will monotonically decrease.
In that case, before each row converge, the diagonal element is always the largest, align with the second law of thermodynamics.
That leads to all rows becomes linear independent, causing the matrix has large rank.
As $t$ increases, some rows will converge, therefore the rank will drop.
% https://zh.wikipedia.org/wiki/%E7%9F%A9%E9%98%B5%E6%8C%87%E6%95%B0

\qed

\end{document}